\documentclass[lettersize,journal,review]{IEEEtran}
\usepackage{algorithm}
\usepackage{amsmath,amsfonts}
\usepackage{array}
 \usepackage{multirow}
\usepackage{booktabs} 
\usepackage[caption=false,font=normalsize,labelfont=sf,textfont=sf]{subfig}
\usepackage{textcomp}
\usepackage{stfloats}
\usepackage{url}
\usepackage{verbatim}
\usepackage{graphicx}
\usepackage{cite}
\usepackage{algpseudocode}
\usepackage{dsfont}

\usepackage[pagebackref,breaklinks,colorlinks]{hyperref}

\makeatletter
\def\abstract{\normalfont
  \if@twocolumn
    \@IEEEabskeysecsize\bfseries\textit{\abstractname}:\space\relax
  \else
    \bgroup\par\addvspace{0.5\baselineskip}\centering\vspace{-1.78ex}\@IEEEabskeysecsize\textbf{\abstractname}\par\addvspace{0.5\baselineskip}\egroup\quotation\@IEEEabskeysecsize
  \fi\@IEEEgobbleleadPARNLSP}
\def\IEEEkeywords{\normalfont
  \if@twocolumn
    \@IEEEabskeysecsize\bfseries\textit{\IEEEkeywordsname}:\space\relax
  \else
    \bgroup\par\addvspace{0.5\baselineskip}\centering\@IEEEabskeysecsize\textbf{\IEEEkeywordsname}\par\addvspace{0.5\baselineskip}\egroup\quotation\@IEEEabskeysecsize
  \fi\@IEEEgobbleleadPARNLSP}
\makeatother

\begin{document}

\definecolor{blue}{rgb}{0,0,0}

\title{SpiderMark: Hybrid Watermarking for Generative AI via Algorithmic Injection and Learning-Based Verification}

\author{Chihcheng~Hsieh,
        Yang~Song,
        Flora~Salim,
        Martin~Bell,
        Jacinto C. Nascimento,
        Catarina~Moreira,
\thanks{Chihcheng Hsieh and Catarina Moreira are with the Data Science Institute, University of Technology Sydney, Sydney, Australia.}
\thanks{Yang Song and Flora Salim are with the School of Computer Science and Engineering, University of New South Wales, Australia.}
\thanks{Martin Bell is with Canva, Sydney, Australia.}
\thanks{Jacinto C. Nascimento is with the University of Lisbon, Lisbon, Portugal.}
\thanks{This paper was partially supported by Trust and Digital Society.}
}

\markboth{}{}


\maketitle

\begin{abstract}
The proliferation of diffusion-based generative models has intensified the need for reliable methods to trace and verify the provenance of AI-generated images. Existing watermarking approaches for diffusion models typically rely on isotropic frequency patterns and handcrafted detection metrics, which are prone to directional vulnerability and threshold instability across diverse image transformations. We propose SpiderMark, a hybrid watermarking framework that combines algorithmic frequency-domain injection with learning-based probabilistic verification, requiring no modifications to pretrained diffusion model weights. SpiderMark embeds a structured spider-web pattern, comprising radial spokes and concentric polygonal rings, into the Fourier representation of the initial latent noise, and trains a lightweight ResNet-18 verifier augmented with auxiliary similarity cues (PSNR and L1) to produce \textcolor{blue}{a continuous watermark confidence score} for watermark detection. \textcolor{blue}{Across nine evaluation conditions, SpiderMark achieves the highest fixed-threshold Accuracy in eight, while retaining strong detection on unseen COCO prompts without verifier retraining.} \textcolor{blue}{CLIP similarity and FID measurements show that these detection gains are accompanied by small measured differences in prompt alignment and distributional image quality.} SpiderMark establishes that structured watermark geometry and learned probabilistic verification can be unified into a single, \textcolor{blue}{practical} framework, bridging the gap between imperceptibility, robustness, and real-world verifiability for generative AI content attribution.
\end{abstract}


\section{Introduction}

The rapid proliferation of diffusion-based generative models \cite{ho2020denoising,rombach2022high} has fundamentally transformed the landscape of digital content creation, enabling the synthesis of photorealistic images from textual prompts with unprecedented fidelity and diversity. Models such as Stable Diffusion, DALL-E, and Midjourney have democratized high-quality image generation, making it accessible to millions of users worldwide. However, this accessibility introduces pressing concerns regarding content provenance, copyright protection, and the potential for misuse through deepfakes and misinformation \cite{liu2025review, gupta2024comprehensive,bird2023typology,Rijsbosch2025watermarking}. As AI-generated imagery becomes perceptually indistinguishable from authentic photographs, the ability to reliably trace and verify the origin of digital content has emerged as a critical societal and technical challenge. \textcolor{blue}{The European Union's AI Act, Article 50(2), further motivates machine-readable marking and detection of AI-generated content by setting transparency obligations for providers of AI systems that generate synthetic content~\cite{euaiact2024}.}

Digital watermarking has long served as the principal mechanism for asserting ownership and authenticity in multimedia content ~\cite{barni2001improved,cox2008digital}. Classical approaches embed imperceptible signatures into images by manipulating frequency-domain coefficients, most commonly through the Discrete Wavelet Transform (DWT) ~\cite{raval2003discrete,tian2023multi} or the Discrete Cosine Transform (DCT)~\cite{shensa2002discrete}, and verify their presence through handcrafted correlation or distance metrics. More recently, deep learning has given rise to end-to-end watermarking frameworks, exemplified by HiDDeN~\cite{zhu2018hidden}, which jointly train encoder–decoder architectures with differentiable noise layers to achieve improved robustness against common image distortions. More fundamentally, these post-hoc watermarking methods are applied \emph{after} image generation, rendering them detachable from the generative process and vulnerable to adversarial removal or jailbreaking~\cite{zhao2024invisible}.

 In the context of generative models, a particularly promising paradigm has emerged: in-generation watermarking, wherein the watermark is embedded directly into the generative process rather than applied as a post-hoc modification. This strategy preserves the natural output distribution of the model while ensuring that the watermark propagates through the entire generation pipeline. Recent advances have sought to integrate watermarking directly into the diffusion generation pipeline. Tree-Ring Watermarking~\cite{wen2023tree_ring} embeds concentric ring patterns into the Fourier space of the initial noise vector, producing watermarks that are invisible and structurally invariant to several geometric transformations. The Stable Signature~\cite{fernandez2023stable} fine-tunes the latent decoder of diffusion models to embed a recoverable binary signature into all generated outputs. More recently, methods such as ZoDiac~\cite{zhang2024attack}, Gaussian Shading~\cite{yang2024gaussian}, and ROBIN~\cite{huang2024robin} have introduced further innovations in latent-space watermark injection, adversarial concealment, and distribution-preserving embedding.

 Despite these advances, existing methods exhibit important limitations that hinder their practical deployment. First, approaches that rely on handcrafted detection metrics, such as frequency-domain correlation or distance thresholds, suffer from threshold instability: the optimal decision threshold varies substantially across different image transformations (e.g., JPEG compression, geometric warping, cropping), rendering a single fixed threshold unreliable for real-world verification. Second, most in-generation watermarking methods employ isotropic frequency structures (i.e. concentric rings, uniform grids, or random masks) that distribute watermark energy uniformly by radius. Such patterns are inherently vulnerable to directional distortions, including motion blur, anisotropic resampling, and compression artefacts that preferentially attenuate specific orientations. Third, the verification strategies used by current methods typically operate in a binary, threshold-dependent manner and do not produce \textcolor{blue}{learned continuous confidence scores}, limiting their utility in deployment scenarios that require confidence-aware decision-making under heterogeneous and unknown distortion conditions.

 In this paper, we propose SpiderMark, a hybrid of frequency-domain watermark injection and learned verification for diffusion-generated images. \textcolor{blue}{Injection modifies the initial latent without retraining the generator; verification uses a compatible pretrained model for inversion.} The framework has three components. First, a spider-web pattern with radial spokes and concentric polygonal rings is inserted into the Fourier representation of the initial latent, \textcolor{blue}{providing angular as well as radial spectral structure}. Second, matched clean and watermarked generations under the same prompts and seeds supply verifier training data. Third, a ResNet-18 verifier combines recovered spectral features with PSNR and L1 similarity cues to output \textcolor{blue}{a continuous watermark confidence score}. \textcolor{blue}{The same injection construction is tested on SD2.1 and SDXL; the verifier can be adapted when the generator changes.}

 The main contributions of this work are as follows:

\begin{itemize}
    \item \textbf{A novel hybrid watermarking framework} that unifies algorithmic frequency-domain injection with learning-based probabilistic verification in a single pipeline. \textcolor{blue}{Injection does not modify pretrained diffusion-model weights; verification uses a compatible generator and may require verifier adaptation across architectures.}

    \item \textbf{The spider-web watermark pattern}, a geometrically structured frequency-domain design that introduces both radial and angular components. This interconnected lattice supplies directional information, spatial redundancy, and balanced frequency coverage; \textcolor{blue}{the complete SpiderMark system empirically outperforms the evaluated ring-based and grid-based systems under anisotropic and localised distortions.}

    \item \textbf{A learned frequency-domain verifier} based on a fine-tuned ResNet-18 backbone that fuses deep spectral features with explicit similarity cues (PSNR and L1 distance) computed against the known watermark pattern. The verifier operates on the Fourier spectrum of the latent noise recovered via inverse DDIM, outputting \textcolor{blue}{a continuous confidence score that enables stable watermark detection under a fixed decision threshold. It does not require the original prompts, noise seeds, or generation session; deterministic DDIM inversion uses a compatible copy of the corresponding pretrained diffusion model and sampling schedule.}
\end{itemize}

These contributions establish SpiderMark as a practical and generalisable solution for content attribution in diffusion-based generative models, addressing key limitations of existing approaches in robustness, verification reliability, and cross-dataset transferability. The rest of this paper is organized as follows. Section 2 reviews related work. Section 3 details the SpiderMark methodology, including watermark injection, spider-web pattern design, and the learned verification pipeline. Section 4 presents experimental results and ablation studies. Section 5 discusses implications for robustness evaluation and deployment. Section 6 concludes the paper and outlines future directions. 

\section{Related Work}
\label{sec:rel_work}

Watermarking AI-generated content has become a critical area of research due to the rapid 
adoption of generative models and the concomitant concerns around provenance, copyright 
protection, and content authenticity. The proliferation of hyper-realistic synthetic media, 
including DeepFakes, has intensified the need for reliable attribution and verification 
mechanisms~\cite{gupta2024comprehensive, nguyen2022deep, verdoliva2020media}. 
Recent work spans techniques that embed imperceptible signatures into the image generation 
process itself, frequency-domain watermarking methods, and detection approaches that 
leverage deep learning for content authentication. These classes of methodologies are 
described next.

\noindent
\textbf{Content Authentication in the Era of Generative AI.}
The growing sophistication of generative models has rendered traditional detection 
approaches increasingly insufficient. As surveyed by Gupta~et~al.~\cite{gupta2024comprehensive}, 
DeepFake detection methods have evolved from simple spatial artifact analysis to 
multi-modal fusion techniques combining spatial, temporal, and frequency-domain 
features. Frequency-based approaches have proven particularly effective: 
Neves~et~al.~\cite{neves2020ganprintr} demonstrated that the Discrete Fourier 
Transform can expose statistical artifacts in GAN-generated frames, compressing 
2D amplitude spectra into feature vectors for classification. 
Gu{\"e}ra and Delp~\cite{guera2018deepfake} showed that temporal inconsistencies 
introduced by face-swapping can be captured through CNN--RNN pipelines operating 
on frame-level features.
However, detection-based approaches are fundamentally reactive: they attempt to identify 
synthetic content after generation, without any cooperation from the generative model. 
This motivates \emph{proactive} approaches based on digital watermarking, which embed 
verifiable signatures during or after the generation process, enabling content attribution 
regardless of downstream analysis capabilities.

\noindent
\textbf{Embedded Watermarking in Generative Models.}
Rezaei~et~al.~\cite{rezaei2024lawa} propose LaWa, a watermarking framework that 
leverages the latent space of latent diffusion models (LDMs) to integrate watermarks 
during generation, achieving robustness across transformations while maintaining 
perceptual quality. This approach embeds signatures directly into generative pipelines 
rather than performing post-hoc modifications, demonstrating improved resilience and 
low false positive rates compared to earlier methods. A similar line of work focuses 
on embedding watermarks specifically within diffusion processes. 
Chen~et~al.~\cite{chen2025tagwm} introduce Dynamic Watermarks for diffusion models, 
using multi-stage embedding that includes a fixed watermark in the model's learned noise 
distribution and a dynamic, image-adapted watermark decoded via a fine-tuned neural 
decoder. This enables robust source verification with minimal visual impact.

Tree-Ring Watermarking~\cite{wen2023tree_ring} embeds concentric ring patterns into 
the Fourier space of the initial noise vector, producing watermarks that are invisible 
and structurally invariant to several geometric transformations. 
The Stable Signature~\cite{fernandez2023stable} fine-tunes the latent decoder of 
diffusion models to embed a recoverable binary signature into all generated outputs. 
More recently, methods such as ZoDiac~\cite{zhang2024attack}, 
Gaussian Shading~\cite{yang2024gaussian}, and ROBIN~\cite{huang2024robin} have 
introduced further innovations in latent-space watermark injection, adversarial concealment, and distribution-preserving embedding.

Despite these advances, existing in-generation methods exhibit important limitations. 
First, approaches relying on handcrafted detection metrics such as frequency-domain 
correlation or distance thresholds suffer from \emph{threshold instability}: the optimal 
decision boundary varies substantially across different image transformations. 
Second, most methods employ isotropic frequency structures (concentric rings, uniform 
grids, or random masks) that distribute watermark energy uniformly by radius, making 
them vulnerable to directional distortions including motion blur and anisotropic 
resampling. Third, existing verification strategies typically operate in a binary, 
threshold-dependent manner without producing \textcolor{blue}{learned continuous confidence scores}.

\noindent\textbf{Frequency-Domain Watermarking: Fourier and Wavelet Transforms.}
The Discrete Fourier Transform (DFT) has long been a fundamental tool in signal and 
image processing. Several works~\cite{jia2023fourier, jiang2021focal, rao2021global, 
yu2022deep} have explored the use of DFT for image analysis, exploiting 
frequency-domain representations to enhance model performance and gain deeper 
insights into image characteristics. For instance, Baig~\cite{baig2022dft} employs 
DFT to globally assess the quality of blurred images, while 
Rao~\cite{rao2021global} uses frequency analysis to capture global image properties. 
These studies demonstrate that DFT is particularly effective for extracting global 
information from images, motivating its use in watermark injection and verification.

Concerning DWT-based approaches, the Discrete Wavelet Transform processes signals 
by decomposing them into multiple frequency bands at varying spatial resolutions. 
This multi-scale decomposition makes DWT especially well suited for capturing 
localized details. Prior studies~\cite{mohideen2008image, pimpalkhute2021digital} have 
successfully applied DWT to image denoising tasks. In particular, 
Tian~\cite{tian2023multi} highlights the benefit of DWT in jointly representing 
spatial and frequency information through multi-resolution analysis.

In the context of watermarking, classical approaches~\cite{barni2001improved,raval2003discrete,shensa2002discrete,tao2004robust} often exploit DWT subbands for embedding 
data. However, these methods were designed for natural images and do not account for 
the unique properties of diffusion-generated content, where the watermark can be 
injected directly into the generative noise.

\noindent\textbf{Deep Learning for Watermarking and Steganography.}
Digital watermarking is a technique for safeguarding digital content by asserting 
copyright ownership. Its key distinction from steganography lies in the embedding 
priority: watermarking focuses on robustness, ensuring the embedded information 
remains detectable even after modifications or distortions, whereas steganography 
emphasizes invisibility. With the recent advances in deep learning, a plethora of 
learned watermarking methods have emerged. To the best of our knowledge, the 
HiDDeN~\cite{zhu2018hidden} approach was the first end-to-end deep watermarking 
framework that embeds resilient messages via a differentiable noise layer. 
More recently, Zhao~et~al.~\cite{zhao2024invisible} demonstrated that invisible 
image watermarks are provably removable using generative AI, underscoring the 
vulnerability of post-hoc watermarking methods and motivating in-generation 
approaches such as SpiderMark.

\noindent\textbf{\textcolor{blue}{Adjacent Post-Hoc and Authentication-Oriented Watermarking.}}
{\color{blue}
Recent methods also combine learned detection, frequency-domain embedding,
and authentication in conventional image-watermarking settings. Hacini
et~al.~\cite{hacini2026neural} propose neural-guided spread-spectrum
watermarking for diagnostic medical images, using learned perceptual masking,
adaptive frequency-subband selection, and a neural detector. SEF-QIM
\cite{dhar2026sefqim} derives authentication features using SVD and QR
decomposition and embeds them through quantization index modulation for tamper
detection and localization. Sahu and Sahu~\cite{sahu2026hybridfragile} combine
DWT, DCT, and SVD embedding with tamper localization and dual self-recovery.

These methods advance robust embedding, learned verification, and image
authentication, but watermark already existing images and target objectives
such as medical-image integrity, tamper localization, or content recovery.
Quantum watermarking has also been explored through NEQR image representation
and reversible quantum logic in simulation~\cite{sahu2026quantum}. Its image
representation and computational setting differ fundamentally from
diffusion-based generation. SpiderMark instead addresses provenance at the
source: it embeds a structured watermark in the initial noise of a diffusion
generation process and detects the propagated signature after downstream
transformations. Consequently, direct numerical comparison with these four
methods would mix different tasks, embedding stages, threat models, and
metrics. Our quantitative comparisons focus on in-generation diffusion
watermarking methods, including Gaussian Shading and ROBIN, evaluated under a
common nine-condition protocol (Section~\ref{subsec:recent_method_comparison}).
}

\noindent\textbf{Positioning of SpiderMark.}
While many recent contributions embed watermarks within generative models, most focus 
on latent representation manipulation (e.g., LaWa), dynamic multi-stage embedding 
(e.g., Dynamic Watermarks), or distribution-preserving strategies (e.g., Gaussian 
Shading) for diffusion models, SpiderMark differentiates itself by proposing a truly 
hybrid watermarking framework that couples algorithmic frequency-domain injection with 
learning-based probabilistic verification. Unlike prior approaches that rely on 
handcrafted metrics or isotropic frequency patterns, SpiderMark introduces a structured 
spider-web pattern providing directional robustness and trains a lightweight verifier 
that outputs \textcolor{blue}{continuous watermark confidence scores}, addressing the threshold instability that 
limits existing methods in real-world deployment scenarios.

\section{Methodology}

\begin{figure*}
    \centering
    \includegraphics[width=.9\linewidth]{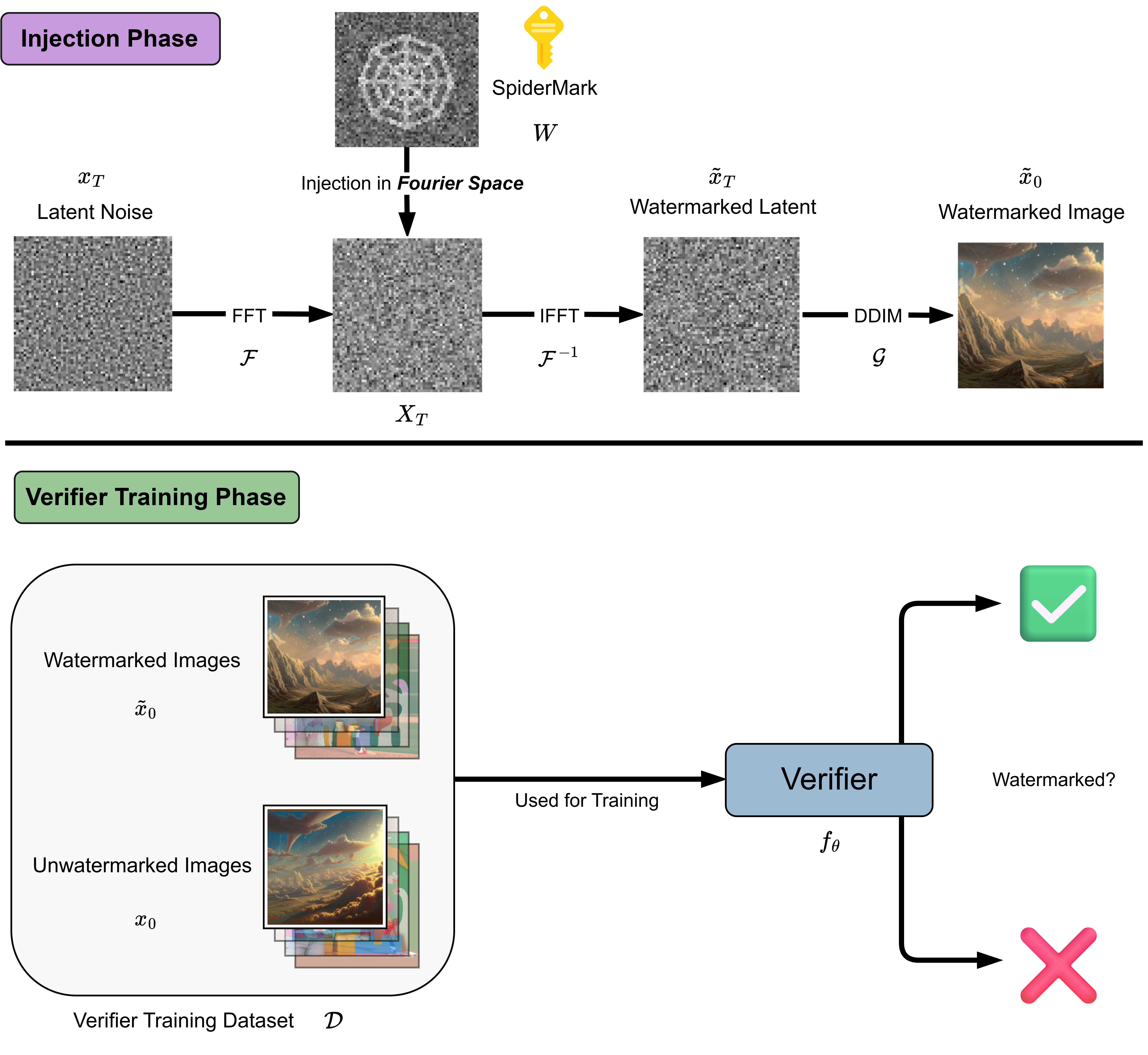}
    \caption{Overview of the proposed SpiderMark watermarking framework.
\textbf{Top}: Injection phase, where a fixed SpiderMark pattern is injected into the initial diffusion noise in the Fourier domain before DDIM sampling, producing a watermarked image without modifying the pretrained model.
\textbf{Bottom}: Verifier training and inference, where paired watermarked and unwatermarked images are used to train a verifier that outputs a probabilistic watermark confidence at test time.}
    \label{fig:spidernet}
\end{figure*}

We propose \textit{SpiderMark}, a frequency-domain watermarking framework for diffusion-based image generation that operates entirely at the noise initialisation stage and requires no modification to pretrained model weights. 
The method consists of three tightly coupled components: (i) direct watermark injection into the initial diffusion noise, (ii) verifier dataset construction via paired watermarked and unwatermarked generations, and (iii) probabilistic watermark verification using a lightweight classifier. 
The complete pipeline is summarised and illustrated in Fig.~\ref{fig:spidernet}.

Given a pretrained diffusion model $\mathcal{G}$, SpiderMark injects a fixed, structured watermark pattern into the frequency representation of the initial latent noise prior to the denoising process. 
\textcolor{blue}{Because injection changes only the initial latent, it does not require retraining the generator. We evaluate this construction on SD2.1 and SDXL; verification still requires a compatible model for inversion and may require an adapted verifier.}
To enable robust detection, we generate matched pairs of watermarked and clean images under identical prompts and noise seeds, from which frequency-domain features are extracted to train a verifier network.
At inference time, the verifier outputs \textcolor{blue}{a continuous watermark confidence score} indicating the presence of the watermark in a given image.

In what follows, we first describe the watermark injection mechanism and then detail the verifier's training and inference procedure.

\subsection{Injection Phase}
Let \(x_T \sim \mathcal{N}(0,I)\) denote the initial Gaussian latent used to initialise the diffusion process.
{\color{blue}SpiderMark embeds a structured watermark \(W\) within a predefined frequency-domain support \(S\) of the initial latent \(x_T\), where \(S(u,v)\in\{0,1\}\) is a binary mask indicating the coefficients modified during injection.}
The complete injection and watermarked image generation procedure is summarised in Algorithm~\ref{alg:injection_generation}.

The Fourier transform of the latent noise is first computed as
\begin{equation}
    X_T = \mathcal{F}\{x_T\},
\end{equation}
where the transform operates over the two-dimensional spatial dimensions of the latent.
{\color{blue}The watermark pattern is blended with the original spectrum only within this support to balance robustness and imperceptibility:}
\begin{equation}
    \label{eq:watermark_injection}
    {\color{blue}\tilde{X}_T
    = (1-S)\odot X_T
    + S\odot\left[(1-\alpha)X_T+\alpha W\right],}
\end{equation}
{\color{blue}where \(\odot\) denotes element-wise multiplication and \(\alpha \in [0,1]\) controls the watermark embedding strength within the selected support. Coefficients outside \(S\) remain unchanged.}
The modified spectrum is subsequently transformed back to the spatial domain:
\begin{equation}
    \tilde{x}_T = \mathcal{F}^{-1}\{\tilde{X}_T\}.
\end{equation}
The resulting watermarked latent serves as the starting point for the denoising trajectory of the pretrained diffusion model, producing a final image \(\tilde{x}_0\) that implicitly carries the embedded watermark.
\textcolor{blue}{The injection step is lightweight and does not modify pretrained diffusion weights; the verifier's cross-generator transfer is evaluated separately in Section~\ref{subsec:sdxl_generalisation}.}

\begin{algorithm}[t]
\caption{Watermark Injection and Watermarked Image Generation}
\label{alg:injection_generation}
\begin{algorithmic}[1]
\Require
Pretrained diffusion model $\mathcal{G}$;
fixed frequency-domain watermark pattern $W$;
{\color{blue}binary frequency-domain injection support $S$;}
injection strength $\alpha \in [0,1]$;
prompt $\mathbf{p}$.

\Ensure
Watermarked latent noise $\tilde{x}_T$;
watermarked generated image $\tilde{x}_0$.

\State Sample initial latent noise $x_T \sim \mathcal{N}(0,I)$.
\State Compute Fourier representation $X_T \leftarrow \mathcal{F}\{x_T\}$.
\State Inject watermark in the frequency domain:
\State \hspace{1.2em} ${\color{blue}\tilde{X}_T \leftarrow (1-S)\odot X_T + S\odot\left[(1-\alpha)X_T+\alpha W\right]}$.
\State Invert transform: $\tilde{x}_T \leftarrow \mathcal{F}^{-1}\{\tilde{X}_T\}$.
\State Generate watermarked image: $\tilde{x}_0 \leftarrow \mathcal{G}(\tilde{x}_T, \mathbf{p})$.
\end{algorithmic}
\end{algorithm}

\paragraph{Spider-web Pattern Generation.}
The watermark \(W(u,v)\) is generated as a regular spider-web pattern that consists of \(n_s\) evenly spaced radial spokes and \(n_\ell\) polygonal ring connectors.  
Let the image dimensions be \(H \times W\), and let the center coordinates be \((x_c,y_c)=(W/2,H/2)\).  
For each spoke \(i \in \{1,\ldots,n_s\}\), we define the angle
\begin{equation}
    \theta_i = \frac{2\pi(i-1)}{n_s},
\end{equation}
and for each ring level \(k \in \{1,\ldots,n_\ell\}\) we define a normalized radius
\begin{equation}
    r_k = r_{\max}\!\left(\frac{k}{n_\ell}\right), \quad r_{\max} = \tfrac{1}{2}\min(H,W)\,\rho,
\end{equation}
where \(\rho\in(0,1)\) determines the overall size of the web relative to the canvas.  
The coordinates of each web node are
\begin{equation}
    (x_{i,k},y_{i,k}) = (x_c + r_k\cos\theta_i,\; y_c + r_k\sin\theta_i).
\end{equation}
Each spoke is represented as a line segment from \((x_c,y_c)\) to \((x_{i,n_\ell},y_{i,n_\ell})\),  
and each ring is formed by connecting adjacent spokes:
\begin{equation}
    (x_{i,k},y_{i,k}) \longleftrightarrow (x_{i+1,k},y_{i+1,k}),
\end{equation}
with \(x_{n_s+1,k}=x_{1,k}\) to close the loop.  
The resulting binary mask \(M(x,y)\in\{0,1\}\) is normalized as
\begin{equation}\label{eq:mask}
    W(u,v)=\frac{M(x,y)-\mu_M}{\sigma_M},
\end{equation}
to form a zero-mean, unit-variance carrier before frequency-space blending.

\paragraph{Advantages of the Spider-web Pattern.}
Most existing watermarking techniques employ simple geometric or frequency-based designs such as concentric rings, grids, or random masks. These isotropic structures distribute watermark energy uniformly by radius, making them vulnerable to directional distortions such as motion blur, anisotropic resampling, or compression artifacts. See Figure~\ref{fig:spidernet} for illustration of the pattern.

The spider-web pattern introduces both radial and angular components, forming an interconnected lattice of concentric rings and vertical spokes. This structure offers several key advantages:
\begin{itemize}
    \item \textit{Directional robustness:} Energy distributed across vertical, horizontal, and diagonal orientations \textcolor{blue}{provides complementary angular evidence under directionally biased degradations, unlike a circular-only ring pattern.}
    \item \textit{Spatial redundancy:} The spokes connect low- and mid-frequency regions, preserving partial watermark cues even under cropping or localized corruption.
    \item \textit{Frequency balance:} Vertical and diagonal components overlap slightly with low-frequency bands near the spectral center, providing improved robustness without introducing visible global distortion.
\end{itemize}
Together, these properties \textcolor{blue}{motivate an interpretable directional carrier; the controlled sensitivity and full-system comparisons reported below evaluate its empirical robustness and image quality.}

\subsection{\textcolor{blue}{Angular-Frequency Characterisation}}
\label{subsec:angular_characterisation}
{\color{blue}
Let a Fourier-domain watermark be written in polar coordinates as
$W(r,\theta)$. An ideal isotropic ring has the form
$W_{\mathrm{ring}}(r,\theta)=f(r)$ and therefore satisfies
$\partial W_{\mathrm{ring}}/\partial\theta=0$. Equivalently, in the angular
Fourier expansion
\begin{equation}
W(r,\theta)=\sum_{m\in\mathbb{Z}} a_m(r)e^{im\theta},
\end{equation}
the ideal ring is concentrated in the $m=0$ component and carries no explicit
angular information. SpiderMark can instead be expressed as
$W_{\mathrm{spider}}(r,\theta)=f(r)+g(r,\theta)$, where the spokes and
cross-links make $\partial W_{\mathrm{spider}}/\partial\theta\neq0$ and
produce nonzero angular harmonics.

Under a rotation by $\phi$,
\begin{equation}
W^{\phi}(r,\theta)=W(r,\theta-\phi), \qquad
a_m(r)\mapsto a_m(r)e^{-im\phi}.
\end{equation}
Thus the directional components undergo structured phase shifts rather than
collapsing to a purely radial statistic, providing the learned verifier with
explicit angular evidence. This analysis characterises the mechanism and is
not a formal proof of robustness to arbitrary transformations; the empirical
geometry and transformation studies in
Sections~\ref{subsec:geometry_sensitivity} and
\ref{subsec:strong_transformations} evaluate the resulting robustness.
}

\subsection{Verification Phase}

The verification phase determines whether a given image contains a watermark. Traditional verification methods typically rely on handcrafted similarity or distance metrics, such as correlation or spectral energy comparisons~\cite{Potdar2005watermarksurvey, wen2023tree_ring, yang2024gaussian}. While effective on clean images, these approaches degrade substantially under common attacks, including cropping and geometric distortions. SpiderMark instead adopts a learned verifier operating in the frequency domain, designed to detect the subtle and structured spectral signatures induced by the injected watermark. {\color{blue}The reported verifier receives both L1 distance and PSNR as auxiliary inputs. Section~\ref{subsec:verifier_ablation} evaluates the combined contribution of these cues while keeping watermark injection fixed; it does not attribute a separate gain to either cue.} This learned formulation generalises well to unseen perturbations and maintains strong detection performance under real-world image transformations.

\paragraph{Verifier Training Dataset ($\mathcal{D}$) Construction}

\begin{algorithm}[t]
\caption{Verifier Training Dataset Construction}
\label{alg:verifier_dataset}
\begin{algorithmic}[1]
\Require
Pretrained diffusion model $\mathcal{G}$;
frequency-domain watermark pattern $W$;
{\color{blue}binary frequency-domain injection support $S$;}
injection strength $\alpha$;
prompt set $\mathcal{P}$;
number of prompts $N$.

\Ensure
Verifier training set $\mathcal{D}$ of labeled images.

\State Initialize dataset $\mathcal{D} \leftarrow \emptyset$.
\For{$i = 1$ to $N$}
    \State Sample prompt $\mathbf{p}_i \sim \mathcal{P}$ and latent noise $x_T^{(i)} \sim \mathcal{N}(0,I)$.
    \State Generate clean image: $x_0^{(i)} \leftarrow \mathcal{G}(x_T^{(i)}, \mathbf{p}_i)$.

    \State Compute spectrum: $X_T^{(i)} \leftarrow \mathcal{F}\{x_T^{(i)}\}$.
    \State Inject watermark in frequency domain:
    \State \hspace{1.2em} ${\color{blue}\begin{aligned}[t]
        \tilde{X}_T^{(i)} \leftarrow {} & (1-S)\odot X_T^{(i)} \\
        & {}+ S\odot\left[(1-\alpha)X_T^{(i)}+\alpha W\right]
    \end{aligned}}$.
    \State Invert transform: $\tilde{x}_T^{(i)} \leftarrow \mathcal{F}^{-1}\{\tilde{X}_T^{(i)}\}$.
    \State Generate watermarked image: $\tilde{x}_0^{(i)} \leftarrow \mathcal{G}(\tilde{x}_T^{(i)}, \mathbf{p}_i)$.

    \State Add labeled image samples:
    \State \hspace{1.2em} $\mathcal{D} \leftarrow \mathcal{D} \cup \{(x_0^{(i)},0), (\tilde{x}_0^{(i)},1)\}$.
\EndFor
\end{algorithmic}
\end{algorithm}

To train the watermark verifier, we construct a paired dataset of clean and watermarked images generated under identical prompts and latent noise initialisations. 
This controlled pairing ensures that the verifier learns to detect watermark-induced spectral signatures, rather than spurious differences arising from semantic content or prompt variation.

Specifically, for each sampled prompt, we draw a single latent noise realisation and use it to generate two images: a clean image produced from the original noise, and a watermarked image generated from the corresponding noise after frequency-domain watermark injection. 
Both images are produced using the same pretrained diffusion model, guaranteeing consistency across all other factors.

The resulting dataset consists of labeled image pairs, where clean images are assigned label~$0$ and watermarked images are assigned label~$1$. Algorithm~\ref{alg:verifier_dataset} summarises the verifier dataset construction procedure.

\begin{figure*}
    \centering
    \includegraphics[width=1\linewidth]{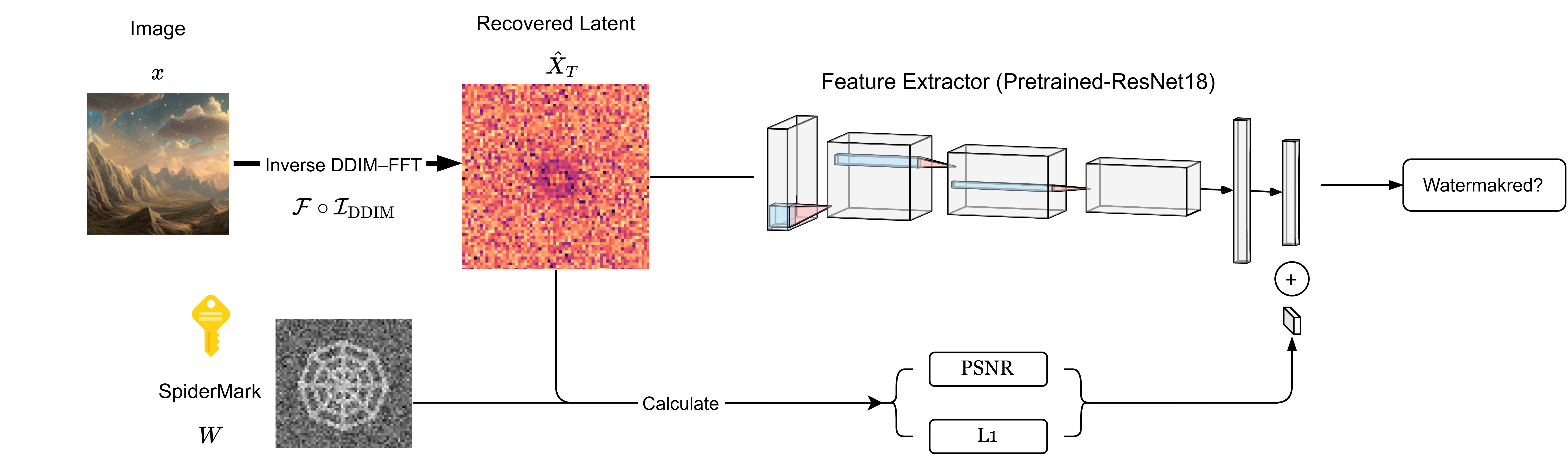}
    \caption{Verifier ($f_\theta$) Architecture. Given an input image ($x$), we first recover its corresponding watermark latent via inverse DDIM ($\mathcal{I}_{\text{DDIM}}$) followed by a Fourier transform ($\mathcal{F}$). The resulting frequency-domain representation is processed by a pretrained ResNet-18 verifier, augmented with auxiliary similarity cues (PSNR and L1) computed against the fixed SpiderMark pattern ($W$). The verifier outputs a probabilistic prediction indicating whether the image is watermarked.}
    \label{fig:veifier_arch}
\end{figure*}

\paragraph{Verifier Training}
The verification network $f_{\theta}$ is trained as a binary classifier to distinguish between watermarked and non-watermarked images.
An overview of the training phase is shown in the bottom part of Fig.~\ref{fig:spidernet}, and the verifier architecture is illustrated in Fig.~\ref{fig:veifier_arch}.
The complete training procedure is summarised in Algorithm~\ref{alg:verifier_training}.

The network is optimised using the binary cross-entropy loss
\begin{equation}
    \mathcal{L} = - \frac{1}{N} \sum_{i=1}^{N}
    \left[y_i \log f_{\theta}(x_i) + (1 - y_i)\log(1 - f_{\theta}(x_i))\right],
\end{equation}
where $y_i \in \{0,1\}$ indicates whether a watermark is present.
The verifier outputs a confidence score $p_i = f_{\theta}(x_i)$, which is thresholded at $0.5$ to obtain a binary prediction.

By combining learned spectral representations with explicit similarity cues based on $\ell_1$ distance and PSNR, SpiderMark leverages both data-driven feature learning and handcrafted watermark-aligned measures.

\begin{algorithm}[t]
\caption{Verifier Training}
\label{alg:verifier_training}
\begin{algorithmic}[1]
\Require
Verifier network $f_\theta$;
loss function $\ell$ (binary cross-entropy);
optimiser \textsc{Opt};
epochs $E$;
batch size $B$;
training set $\mathcal{D} = \{(\textbf{x}^{(i)}, y^{(i)})\}_{i=1}^{2N}$,
where $\textbf{x}^{(i)} \in \{x_0^{(k)}, \tilde{x}_0^{(k)}\}$ and $y^{(i)} \in \{0,1\}$.

\Ensure
Trained verifier parameters $\theta$.

\For{$e = 1$ to $E$}
    \For{each mini-batch $\mathcal{B} = \{(\textbf{x}^{(j)}, y^{(j)})\}_{j=1}^{B}$ sampled from $\mathcal{D}$}
        \State Compute predictions: $\hat{y}^{(j)} \leftarrow f_\theta(\textbf{x}^{(j)})$ for all $(\textbf{x}^{(j)},y^{(j)}) \in \mathcal{B}$.
        \State Compute loss: $\mathcal{L} \leftarrow \frac{1}{B}\sum_{j=1}^{B}\ell(\hat{y}^{(j)}, y^{(j)})$.
        \State Update parameters: $\theta \leftarrow \textsc{Opt}(\theta, \nabla_\theta \mathcal{L})$.
    \EndFor
\EndFor
\end{algorithmic}
\end{algorithm}

\paragraph{Watermark Verification (Inference)}

Given a query image $x$, SpiderMark determines the presence of the embedded watermark through a lightweight inference procedure that operates in the frequency domain.
\textcolor{blue}{Unlike injection and dataset construction, verification does not require the original prompts, noise seeds, or generation session. It does require a compatible copy of the corresponding pretrained diffusion model and sampling schedule to perform deterministic inverse DDIM.}

For verification, we apply the inverse DDIM procedure corresponding to the same pretrained diffusion model and noise schedule used during generation.
We treat inverse DDIM as a deterministic operator that approximately recovers the terminal latent noise from a generated image.
Let $\mathcal{I}_{\text{DDIM}}(\cdot)$ denote this inverse mapping, and compute
\begin{equation}
    \hat{x}_T = \mathcal{I}_{\text{DDIM}}(x).
\end{equation}

The recovered latent $\hat{x}_T$ is transformed into the frequency domain via a two-dimensional Fourier transform,
\begin{equation}
    \hat{X}_T = \mathcal{F}\{\hat{x}_T\},
\end{equation}

Since SpiderMark embeds a fixed and known watermark pattern $W$ during injection, we additionally compute explicit similarity measures between the recovered spectrum and the reference pattern within the watermark support.
Let $M(u,v)$ denote a binary mask defining the watermark support in the frequency domain (as shown in Equation~\ref{eq:mask}).
We compute a masked $\ell_1$ distance
\begin{equation}
    d_{\ell_1} = \frac{1}{\|M\|_1}\left\| M \odot \left(\hat{X}_T - W\right) \right\|_1,
\end{equation}
and a masked peak signal-to-noise ratio (PSNR),
\begin{equation}
    \mathrm{PSNR}_M = 10 \log_{10}\!\left( \frac{\mathrm{MAX}^2}{\mathrm{MSE}_M + \epsilon} \right),
\end{equation}
where
\begin{equation}
    \mathrm{MSE}_M = \frac{1}{\|M\|_1}\left\| M \odot \left(\hat{X}_T - W\right) \right\|_2^2.
\end{equation}
Here, $\mathrm{MAX}$ denotes the maximum possible value of the normalized real-valued representation of $\hat{X}_T$, and $\epsilon$ is a small constant for numerical stability.

Finally, the recovered spectrum $\hat{X}_T$ is processed by the verifier network $f_\theta(\cdot)$, which internally extracts deep frequency-features and fuses them with the two scalar similarity cues $(\mathrm{PSNR}_M, d_{\ell_1})$ in its final prediction layer.
Concretely, the backbone of $f_\theta$ produces an embedding $h$, which is concatenated with $(\mathrm{PSNR}_M, d_{\ell_1})$ and fed into the last layer to output \textcolor{blue}{a continuous watermark confidence score} $p \in [0,1]$.
A binary verification decision is obtained by thresholding $p$ at a fixed operating point $\tau$.
The complete inference procedure is summarised in Algorithm~\ref{alg:verification_inference}.

\begin{algorithm}[t]
\caption{Watermark Verification (Inference)}
\label{alg:verification_inference}
\begin{algorithmic}[1]
\Require
Query image $x$;
inverse DDIM operator $\mathcal{I}_{\text{DDIM}}(\cdot)$;
Fourier transform $\mathcal{F}\{\cdot\}$;
reference watermark pattern $W$ and mask $M$;
trained verifier $f_\theta(\cdot)$;
normalization constant $\mathrm{MAX}$;
stability constant $\epsilon$;
decision threshold $\tau$.

\Ensure
Watermark confidence $p \in [0,1]$;
verification decision $\mathds{1}[p \ge \tau]$.

\State Recover terminal latent noise: $\hat{x}_T \leftarrow \mathcal{I}_{\text{DDIM}}(x)$.
\State Compute Fourier spectrum: $\hat{X}_T \leftarrow \mathcal{F}\{\hat{x}_T\}$.

\State Compute masked $\ell_1$ similarity:
\State \hspace{1.2em} $d_{\ell_1} \leftarrow \frac{1}{\|M\|_1}\left\| M \odot (\hat{X}_T - W) \right\|_1$.
\State Compute masked PSNR:
\State \hspace{1.2em} $\mathrm{MSE}_M \leftarrow \frac{1}{\|M\|_1}\left\| M \odot (\hat{X}_T - W) \right\|_2^2$.
\State \hspace{1.2em} $\mathrm{PSNR}_M \leftarrow 10 \log_{10}\!\left(\frac{\mathrm{MAX}^2}{\mathrm{MSE}_M + \epsilon}\right)$.

\State Predict watermark confidence (fusion occurs in the last layer of $f_\theta$):
\State \hspace{1.2em} $p \leftarrow f_\theta(\hat{X}_T, \mathrm{PSNR}_M, d_{\ell_1})$.

\State \Comment{$f_\theta$ internally computes $h=\phi(\hat{X}_T)$ and fuses
$(\mathrm{PSNR}_M,d_{\ell_1})$ by concatenation in the final prediction layer to output $p \in [0,1]$.}
\State Output decision: $\mathds{1}[p \ge \tau]$.
\end{algorithmic}
\end{algorithm}

\section{Results}

{\color{blue}We first describe the common evaluation protocol and the original
benchmark. We then compare recent in-generation methods and the statistical
support for their observed differences, isolate verifier components,
and examine parameter sensitivity, stronger attacks, cross-generator transfer, and
computational cost. The reference verifier checkpoint is used for the main
benchmark; controlled retraining studies use separate checkpoints, so their
absolute scores should be compared within their respective tables.}

\subsection{Experimental Setup}
SpiderMark is evaluated on a diverse suite of image perturbations to assess watermark detectability and robustness under real-world transformations. 
Experiments are conducted using a Stable Diffusion v2.1 backbone~\cite{rombach2022stable}, where the watermark is injected into the initial noise tensor \(x_T\). 
The verification model is trained to classify between watermarked and non-watermarked images in the frequency domain, and the prediction thresholds $\tau$ for verifier is set to $0.5$.

Nine types of image transformations are applied:  
(1) \textit{clean}, (2) \textit{jpeg\_strong}, (3) \textit{msg\_app\_combo}, (4) \textit{down\_up}, (5) \textit{blur}, (6) \textit{random\_crop}, (7) \textit{occlusion}, (8) \textit{geom\_warp}, and (9) \textit{train\_aug\_mix}.  
These augmentations simulate practical degradations such as compression, scaling, occlusion, and geometric distortion commonly seen in content sharing platforms.

\paragraph{Verifier Training Details.}
We sample prompts from the public \texttt{Gustavosta/Stable-Diffusion-Prompts} collection and generate a balanced dataset of 500 watermarked and 500 non-watermarked images using Stable Diffusion v2.1 \cite{rombach2022stable}.
All images are converted to grayscale, transformed using a 2D FFT, and the magnitude spectrum is provided to a ResNet-18 verifier pretrained on ImageNet.
The model is fine-tuned using binary cross-entropy loss.

\paragraph{\textcolor{blue}{Implementation and Reproducibility Details.}}
{\color{blue}
Unless stated otherwise, images are generated at $512\times512$ with Stable
Diffusion 2.1 Base using 50 denoising steps and classifier-free guidance 7.5.
The reference configuration uses eight spokes, four ring levels, radius ratio
$\rho=0.6$, a centred radius-10 support on the $64\times64$ latent grid, and
$\alpha=0.99$. Generation latents follow seed $42+i$ for image index $i$.
The 500 clean and 500 watermarked images are split 85\%/15\% with split seed
42. The ImageNet-pretrained ResNet-18 verifier is optimized for 300 epochs
with AdamW (learning rate $2\times10^{-4}$, weight decay 0.01), batch size 8,
and binary cross-entropy; the decision threshold is fixed at $\tau=0.5$.
Verification uses 50-step empty-prompt DDIM inversion with guidance scale 1.

The nine-condition evaluator uses JPEG quality sampled from 40--60,
downsample--upsample factor 0.5, random-resized-crop area 0.5--0.9 and aspect
ratio 0.75--1.33, a random black occlusion covering 25\% of each image
dimension, Gaussian-blur radius 1.5--3.0, and a geometric condition combining
rotation within $\pm15^\circ$ with crop area 0.7--1.0. The messaging condition
combines half-resolution scaling, JPEG quality 40--70, and restoration to the
original resolution; the augmentation-mixture parameters follow the verifier
training pipeline. All five reported repetitions reuse fixed checkpoints and
images and vary only seeded stochastic transformations. Experiment scripts
record the prompt/seed rules, checkpoint selection, attack configuration, and
per-run outputs needed to reproduce every added table. Upon publication, we
will release the source code, trained verifier checkpoint, experiment
configurations, and evaluation manifests or dataset-generation instructions
needed to reproduce the study. Third-party images or datasets will be
redistributed only where their licences permit; otherwise, acquisition and
generation instructions will be provided.
}






\begin{figure*}[h!]
    \resizebox{2\columnwidth}{!}{
    \includegraphics{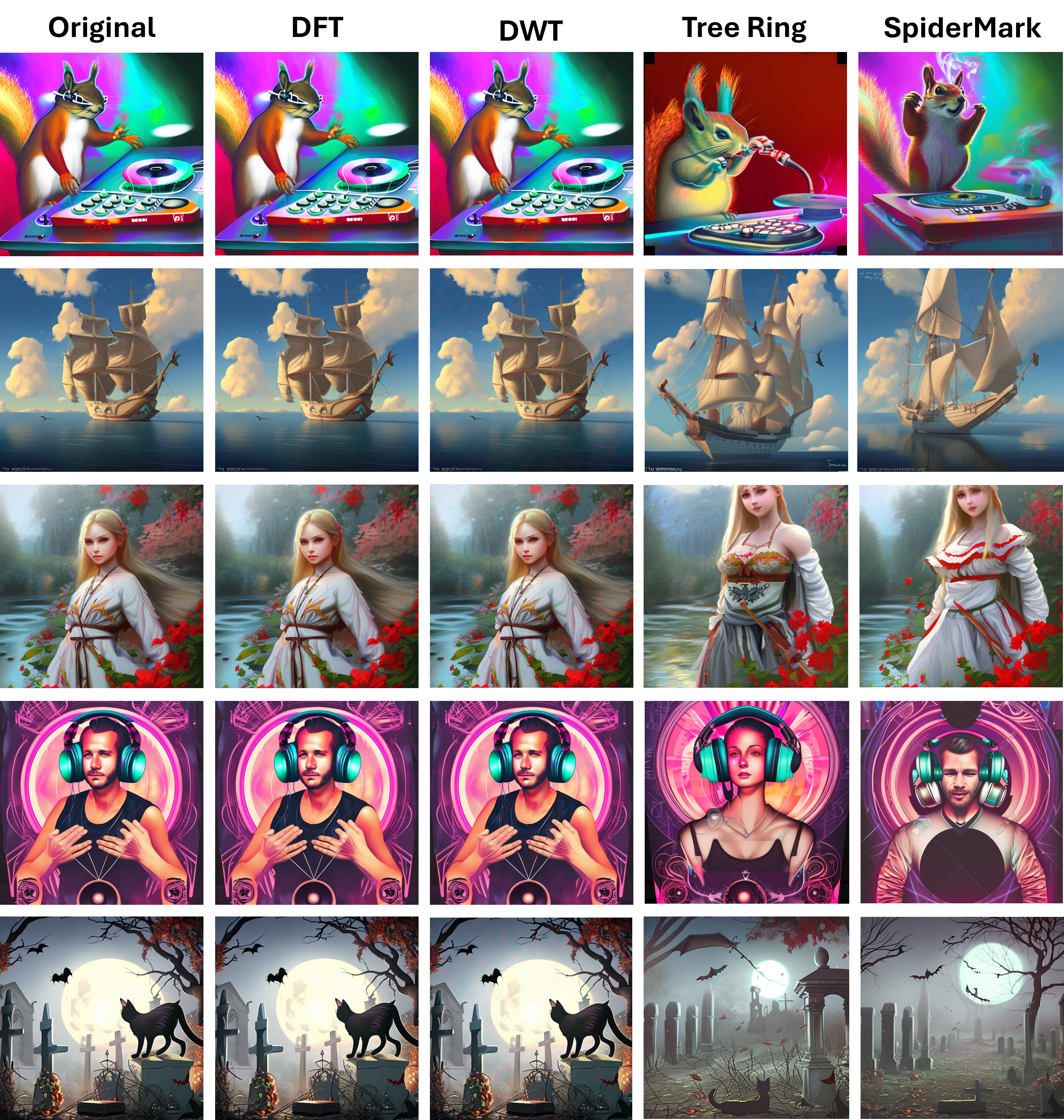}
    }
    \caption{
    \textcolor{blue}{Illustrative outputs from the watermarking pipelines.
    Columns show the original, DFT, DWT--DCT, Tree-Ring, and SpiderMark
    outputs. Generated content differs across some methods, so this panel
    should not be interpreted as a paired image-quality comparison.}
    }
    \label{fig:watermark_comparison}
\end{figure*}

\paragraph{Robustness Evaluation.}
SpiderMark is evaluated under a comprehensive set of post-processing transformations that simulate real-world image distortions. 
Each perturbation is applied to generated images before verification. {\color{blue}Reported repetitions are five seeded evaluation runs with fixed checkpoints; they quantify evaluation variability rather than independent retraining variability.}
The following perturbations are considered:
\begin{itemize}
    \item \textit{Clean}: unmodified images as a reference.
    \item \textit{JPEG-Strong}: aggressive JPEG compression at low quality.
    \item \textit{MsgApp-Combo}: chained resampling and JPEG recompression designed to emulate the typical processing pipeline of messaging applications, where images are resized and compressed before transmission.
    \item \textit{Down-Up}: downsampling followed by upsampling to simulate resolution scaling.
    \item \textit{Blur}: Gaussian blur representing optical or motion blur.
    \item \textit{Random Crop}: random resized crops reflecting framing changes.
    \item \textit{Occlusion}: partial masking to assess redundancy and spatial resilience.
    \item \textit{Geom-Warp}: geometric warping and affine distortion.
    \item \textit{Aug-Mix}: composite augmentations used to test generalisation.
\end{itemize}
Evaluation metrics include accuracy and AUROC, averaged across all trials.

\subsection{Compared Watermarking Approaches}
\label{subsec:compared_methods}

We compare SpiderMark against representative frequency-domain watermarking and verification methods.

\paragraph{Tree-Ring.}
Tree-Ring \cite{wen2023tree_ring} embeds structured circular patterns into the Fourier domain of the initial diffusion noise. 
Verification is performed by measuring the correlation between the extracted frequency spectrum and a predefined ring-shaped template. 
As a diffusion-native watermarking method, Tree-Ring is designed to preserve imperceptibility while maintaining robustness to common image transformations.

\paragraph{DFT-Single.}
DFT-Single embeds watermark signals into selected coefficients of the discrete Fourier transform (DFT) of the image. 
The watermark is injected into specific frequency bands, and verification is typically conducted by measuring reconstruction error or frequency-domain similarity. 
 This approach represents a classical frequency-based watermarking baseline without diffusion-specific design.

\paragraph{DWT--DCT.}
DWT--DCT is a hybrid frequency-domain watermarking method that combines discrete wavelet transform (DWT) and discrete cosine transform (DCT). 
The watermark is embedded into mid-frequency subbands to balance robustness and perceptual quality. 
Verification is performed via coefficient comparison or reconstruction-based similarity metrics.

These baselines cover both diffusion-specific watermarking (Tree-Ring) and conventional frequency-domain watermarking methods (DFT-Single and DWT--DCT), providing a comprehensive comparison against SpiderMark.

\subsection{Quantitative Evaluation}

\begin{table*}[h]
\centering
\caption{\textcolor{blue}{Detection under nine transformations using the reference checkpoints. Bold denotes the highest value in each row; Accuracy uses each method's fixed threshold, while AUROC summarizes score ranking.}}
\label{tab:robustness_results_all}
\resizebox{\textwidth}{!}{
\begin{tabular}{lcccccccc}
\toprule
\multirow{2}{*}{\textbf{Attack}} &
\multicolumn{2}{c}{\textbf{SpiderMark}} &
\multicolumn{2}{c}{\textbf{Tree-Ring~\cite{wen2023tree_ring}}} &
\multicolumn{2}{c}{\textbf{DFT-Single}} &
\multicolumn{2}{c}{\textbf{DWT--DCT}} \\
\cmidrule(lr){2-3}
\cmidrule(lr){4-5}
\cmidrule(lr){6-7}
\cmidrule(lr){8-9}
& Acc & AUROC
& Acc & AUROC
& Acc & AUROC
& Acc & AUROC \\
\midrule

clean           
& 0.9467 & 0.9904
& \textbf{0.9759} & \textbf{0.9971}  
& 0.9090 & 0.9603
& 0.6760 & 0.7199
\\

jpeg\_strong    
& \textbf{0.8720} & \textbf{0.9410}
& 0.6627 & 0.8772
& 0.5717 & 0.5730
& 0.6737 & 0.7156
\\

msg\_app\_combo 
& \textbf{0.6813} & \textbf{0.8468}
& 0.4747 & \underline{0.8136}
& 0.5223 & 0.5044
& 0.6700 & 0.7113
\\

down\_up        
& \textbf{0.9133} & \textbf{0.9713}
& 0.6988 & 0.9576
& 0.5270 & 0.5164
& 0.6720 & 0.7159
\\

blur            
& \textbf{0.8160} & \textbf{0.8757}
& 0.4771 & 0.7777
& 0.5620 & 0.5717
& \underline{0.6690} & 0.7103
\\

random\_crop    
& \textbf{0.8787} & 0.9503
& 0.7807 & \textbf{0.9565}
& 0.5133 & 0.4912
& 0.5220 & 0.5046
\\

occlusion       
& \textbf{0.9574} & 0.9878
& 0.9325 & \textbf{0.9915}
& 0.9080 & 0.9582
& 0.6603 & 0.6939
\\

geom\_warp      
& \textbf{0.8560} & 0.9381
& 0.7711 & \textbf{0.9412}
& 0.5190 & 0.4909
& 0.5193 & 0.5083
\\

train\_aug\_mix 
& \textbf{0.7280} & 0.8620
& 0.6337 & \textbf{0.8634}
& 0.5417 & 0.5340
& 0.5360 & 0.5283
\\

\bottomrule
\end{tabular}

}
\end{table*}

{\color{blue}Table~\ref{tab:robustness_results_all} reports the reference-checkpoint, per-attack comparison. SpiderMark has the highest fixed-threshold Accuracy in eight of nine conditions; Tree-Ring leads on clean images. These are observed detection differences, not per-attack significance claims or image-quality measurements. A separately trained, controlled verifier ablation appears in Table~\ref{tab:verifier_ablation}; its absolute scores are not directly comparable with this benchmark.}

\subsection{Evaluation on COCO Dataset}

To evaluate cross-dataset generalisation, we further assess watermark verification performance on the COCO dataset \cite{lin2014microsoft}, which is not used during verifier training (Algorithm~\ref{alg:verifier_training}). In contrast to prior experiments (Table~\ref{tab:robustness_results_all}) where the verifier is trained and evaluated on images generated from the Stable Diffusion prompt dataset, this setting tests whether the learned verifier overfits to prompt distributions or generation characteristics seen during training.

{\color{blue}Table~\ref{tab:coco_four_methods} shows that SpiderMark retains
the highest reported Accuracy and AUROC across the nine COCO conditions,
including compression, cropping, and geometric warping, without retraining
the verifier on COCO prompts.}

{\color{blue}This supports transfer beyond the prompt set used for training,
although it does not establish generalisation to every dataset or generator.
The COCO result concerns detection; image quality is assessed separately by
the FID row in the same table.}

{\color{blue}SpiderMark's COCO FID is numerically close to the other
watermarked methods, though not the lowest. Taken together, the detection
and quality measurements support prompt-set transfer with a modest measured
quality difference rather than a claim of perceptual equivalence.}

\begin{table*}[h]
\centering
\caption{Watermark verification performance on COCO (unseen dataset) under various image transformations.
FID scores are shown as a separate row for each watermarking method.}
\label{tab:coco_four_methods}

\resizebox{.8\textwidth}{!}{
\begin{tabular}{lcccccccc}
\toprule
\multirow{2}{*}{\textbf{Attack / Metric}} &

\multicolumn{2}{c}{\textbf{SpiderMark}} &
\multicolumn{2}{c}{\textbf{Tree-Ring}} &
\multicolumn{2}{c}{\textbf{DFT-Single}} &
\multicolumn{2}{c}{\textbf{DWT--DCT}} \\

\cmidrule(lr){2-3}
\cmidrule(lr){4-5}
\cmidrule(lr){6-7}
\cmidrule(lr){8-9}

& Acc & AUROC
& Acc & AUROC
& Acc & AUROC
& Acc & AUROC \\
\midrule

clean           
& \textbf{0.9733} & \textbf{0.9925}
& 0.8800 & 0.9483
& 0.8840 & 0.9433
& 0.6630 & 0.7062
\\

jpeg\_strong    
& \textbf{0.9000} & \textbf{0.9735}
& 0.7987 & 0.8800
& 0.5793 & 0.5934
& 0.6590 & 0.7009
\\

msg\_app\_combo 
& \textbf{0.7733} & \textbf{0.8820}
& 0.6360 & 0.7556
& 0.5240 & 0.5106
& 0.6567 & 0.6972
\\

down\_up        
& \textbf{0.9133} & \textbf{0.9720}
& 0.7800 & 0.8992
& 0.5130 & 0.4996
& 0.6600 & 0.7026
\\

blur            
& \textbf{0.8173} & \textbf{0.8877}
& 0.6440 & 0.7694
& 0.5540 & 0.5469
& 0.6557 & 0.6975
\\

random\_crop    
& \textbf{0.8240} & \textbf{0.9169}
& 0.7413 & 0.8451
& 0.5177 & 0.4973
& 0.5180 & 0.5011
\\

occlusion       
& \textbf{0.9720} & \textbf{0.9908}
& 0.8560 & 0.9398
& 0.8820 & 0.9402
& 0.6417 & 0.6832
\\

geom\_warp      
& \textbf{0.8373} & \textbf{0.9300}
& 0.7613 & 0.8377
& 0.5147 & 0.4871
& 0.5317 & 0.5173
\\

train\_aug\_mix 
& \textbf{0.7227} & \textbf{0.8389}
& 0.6813 & 0.7730
& 0.5360 & 0.5330
& 0.5360 & 0.5325
\\

\midrule\midrule
FID score  (non-watermarked = 81.743)
& \multicolumn{2}{c}{\underline{83.2103}}
& \multicolumn{2}{c}{83.5059}
& \multicolumn{2}{c}{\textbf{81.7612}}
& \multicolumn{2}{c}{83.3128} \\
\bottomrule

\end{tabular}
}
\end{table*}

\subsection{Semantic Preservation Evaluation using CLIP}

\begin{table}[t]
\centering
\caption{CLIP image--text semantic similarity comparison between clean (unwatermarked) images and different watermarking methods. Higher values indicate better semantic alignment.}
\label{tab:clip_similarity}
\begin{tabular}{lcc}
\toprule
\textbf{Method} & \textbf{COCO} & \textbf{Stable Diffusion} \\
\midrule
Unwatermarked & \textbf{0.3302} & \textbf{0.3615} \\
Tree-Ring             & 0.3296          & 0.3614          \\
Ours (Spidermark)        & 0.3273          & 0.3599          \\
\bottomrule
\end{tabular}
\end{table}

To assess whether watermark injection alters the semantic content of generated images, we evaluate image--text semantic similarity using CLIP~\cite{radford2021clip}. For each image, we compute the cosine similarity between the CLIP image embedding and its corresponding generation prompt. We report the mean CLIP similarity over all samples, with higher values indicating better semantic alignment.

{\color{blue}Table~\ref{tab:clip_similarity} reports mean CLIP similarity
on COCO and Stable Diffusion prompts. SpiderMark scores 0.3273 and 0.3599,
compared with 0.3302 and 0.3615 for clean images; Tree-Ring scores 0.3296
and 0.3614. The SpiderMark--clean differences are below 0.003 on both
cohorts, indicating that prompt alignment is largely retained under this
metric. These small numerical gaps are not a formal equivalence test or a
standalone perceptual-quality assessment.}

\subsection{\textcolor{blue}{Comparison with Recent In-Generation Methods}}
\label{subsec:recent_method_comparison}
{\color{blue}
To test whether SpiderMark remains competitive against more recent
diffusion-native schemes, we compare it with Gaussian Shading~\cite{yang2024gaussian}
and ROBIN~\cite{huang2024robin}
under the same prompt set and nine-condition evaluation protocol.
The methods use their MarkDiffusion implementations with Stable Diffusion
2.1 Base, the same OctoWeb generation-seed indices, and fixed decision
thresholds without per-attack tuning. Detection uses the 15\% validation
split and five seeded evaluation runs; these runs do not represent
independent retraining. Generation uses guidance 7.5. Empty-prompt inversion
uses guidance 1 for SpiderMark and 7.5 in the recorded baseline detection
runs, unlike the controlled timing protocol in
Section~\ref{subsec:verification_cost}.

At these method-specific operating points, SpiderMark achieves 84.95\%
macro Accuracy, compared with 83.23\% for Gaussian Shading and 78.31\%
for ROBIN (Table~\ref{tab:recent_method_comparison}). This operating point
reduces missed watermarks, although it incurs a higher false-positive rate.
ROBIN achieves the highest AUROC. Thus, the Accuracy advantage is not a
claim of superiority at a matched false-positive rate or across all
thresholds.

On 500 COCO prompts, watermarked FID remains numerically close across
methods: 83.07 for SpiderMark, 82.49 for Gaussian Shading, and 82.87 for
ROBIN. Watermarked prompt CLIP scores span approximately 0.328--0.333.
Here, comparable image quality describes the proximity of the measured
quality metrics, not a statistical equivalence result or a claim that
SpiderMark achieves the lowest FID. DFT-Single and DWT--DCT remain
complementary post-hoc reference baselines rather than direct
in-generation competitors.
}

\begin{table}[t]
\centering
\caption{\textcolor{blue}{Recent in-generation baselines. Accuracy, FPR and
FNR are macro percentages over nine conditions; Accuracy reports mean
$\pm$ SD across five evaluation runs. AUROC is the macro mean.
Watermarked FID is evaluated separately on 500 COCO prompts.}}
\label{tab:recent_method_comparison}
{\color{blue}\setlength{\tabcolsep}{2pt}\small
\begin{tabular}{lccccc}
\toprule
Method & Accuracy $\uparrow$ & AUROC $\uparrow$ & FPR $\downarrow$ & FNR $\downarrow$ & FID $\downarrow$ \\
\midrule
SpiderMark & $84.95\pm0.79$ & 0.9286 & 8.67 & 20.79 & 83.07 \\
Gaussian Shading & $83.23\pm0.20$ & 0.8870 & 0.00 & 31.84 & 82.49 \\
ROBIN & $78.31\pm0.43$ & 0.9663 & 0.06 & 41.13 & 82.87 \\
\bottomrule
\end{tabular}}
\end{table}

{\color{blue}
To assess the strength of these observed differences, we compare
SpiderMark with Gaussian Shading and ROBIN using fixed-threshold Accuracy
across the nine conditions (Table~\ref{tab:recent_method_comparison}).
Table~\ref{tab:statistics} reports the mean Accuracy difference between
five seeded evaluation runs per method, a pointwise 95\% confidence interval,
and a two-sided Welch-test $p$ value adjusted over all 34 evaluated
comparisons by Holm's method. The runs reuse fixed checkpoints and images;
they measure variation in the evaluation procedure, not independent
training or population uncertainty.

SpiderMark's observed advantage over ROBIN is 6.64 percentage points and
remains supported after Holm adjustment. Its 1.72-point advantage over
Gaussian Shading is positive but does not meet the adjusted 0.05 criterion
($p_{\mathrm{Holm}}=0.0737$). With only five runs per method, distributional
assumptions cannot be checked reliably; we therefore treat these tests as
supporting evidence rather than proof of general superiority. The full
34-comparison analysis, including parameter and verifier contrasts and exact
rank-test sensitivity checks, is provided in the supplementary statistical
analysis. A nonsignificant result does not establish equivalence.
}
\begin{table}[t]
\centering
\caption{\textcolor{blue}{Fixed-threshold Accuracy: SpiderMark minus each
comparison method. Pointwise 95\% intervals are
unadjusted; Welch-test $p$ values are Holm-adjusted over 34 evaluated tests.}}
\label{tab:statistics}
{\color{blue}\footnotesize\setlength{\tabcolsep}{3pt}
\begin{tabular}{lccc}
\toprule
Comparison & $\Delta$Accuracy (pp) & 95\% CI & $p_{\mathrm{Holm}}$ \\
\midrule
Gaussian Shading & +1.72 & [0.75, 2.68] & 0.0737 \\
ROBIN & +6.64 & [5.66, 7.61] & $6.27\times10^{-5}$ \\
\bottomrule
\end{tabular}}
\end{table}

\subsection{\textcolor{blue}{Verifier and Similarity-Cue Ablation}}
\label{subsec:verifier_ablation}

{\color{blue}
To separate verifier effects from watermark injection, we hold the SpiderMark
pattern and support fixed and compare three verification configurations
(Table~\ref{tab:verifier_ablation}): classical correlation on the recovered
spectrum, a ResNet-18 using only recovered spectral features, and the full
ResNet-18 with PSNR and L1 similarity cues. The learned variants are trained
separately on the same cached SpiderMark features and evaluated with the same
$\alpha=0.99$ injection, SD2.1 backbone, 150-image validation split, nine
conditions, and five seeded evaluation runs. These runs repeat evaluation,
not training; the controlled retraining results are separate from the
original checkpoint used in the main performance tables.

The full verifier achieves 72.86\% macro Accuracy, compared with 67.88\%
for spectral features alone and 47.33\% for classical correlation. Thus,
the learned verifier substantially improves on the tested handcrafted
detector when the injected pattern is unchanged, and the PSNR--L1 cue pair
provides an additional 4.98 percentage points over the spectral-only
verifier in this controlled protocol. This comparison measures their joint
contribution, not the separate effect of PSNR or L1. Because the pattern is
fixed, it does not establish superiority over alternative geometries;
geometry sensitivity is evaluated separately in
Section~\ref{subsec:geometry_sensitivity}.
}

\begin{table}[t]
\centering
\caption{\textcolor{blue}{Verifier ablations with fixed SpiderMark injection.
Values are macro means $\pm$ SD across five evaluation runs over nine
conditions. The two learned variants are trained separately on the same
cached features; these values are not the reference-checkpoint results.}}
\label{tab:verifier_ablation}
{\color{blue}\footnotesize\setlength{\tabcolsep}{2.5pt}
\begin{tabular}{lcc}
\toprule
Configuration & Accuracy (\%) $\uparrow$ & AUROC $\uparrow$ \\
\midrule
Classical correlation & $47.33\pm0.29$ & $0.4996\pm0.0058$ \\
ResNet-18 (spectral features only) & $67.88\pm0.70$ & $0.7498\pm0.0062$ \\
ResNet-18 + PSNR + L1 (full) & $72.86\pm0.86$ & $0.8261\pm0.0048$ \\
\bottomrule
\end{tabular}}
\end{table}

\subsection{\textcolor{blue}{Sensitivity to Embedding Strength}}
\label{subsec:alpha_sensitivity}
{\color{blue}
To quantify the role of the embedding strength $\alpha$ in
Equation~\ref{eq:watermark_injection}, we evaluate
$\alpha\in\{0.50,0.75,0.99\}$, including the reference setting
$\alpha=0.99$ and two lower strengths.
The diffusion model, prompts, generation seeds, geometry, support and
verifier architecture are held fixed. Each condition uses 500 clean and
500 watermarked SD-prompt images and a separately trained verifier with
five cached augmentation rounds. Checkpoints are selected by validation
loss and evaluated over nine conditions and five seeded evaluation runs
at threshold 0.5.

Table~\ref{tab:alpha_sensitivity} shows that $\alpha=0.99$ achieves the
highest observed fixed-threshold Accuracy, whereas $\alpha=0.75$ achieves
the highest AUROC. Detection performance does not improve monotonically
with strength. On a separate matched 500-prompt COCO cohort, FID increases
from 81.82 to 82.52 to 83.07, compared with 80.25 for the shared clean
reference. Watermarked prompt-CLIP scores remain between 0.3278 and 0.3283
(clean: 0.3296). Paired PSNR is computed as the mean of per-image
$10\log_{10}(255^2/\mathrm{MSE})$ on saved RGB PNGs before any evaluation
attack, without resizing or alignment. Prompts, filenames and clean
pixels match across strengths. PSNR increases slightly, rather than
decreasing, with strength: 14.26--14.50 dB on SD prompts and 12.60--12.77 dB
on COCO prompts. PSNR measures fidelity to the corresponding clean
generation; it does not establish perceptual indistinguishability.
Thus, the results support $\alpha=0.99$ as a reasonable fixed-threshold
setting among those tested, with a numerically modest FID increase, rather
than a universal optimum or a monotonic trade-off in every metric.
}

\begin{table}[t]
\centering
\caption{\textcolor{blue}{Embedding-strength sensitivity. Accuracy and
AUROC are macro means over nine conditions, with SD over five evaluation
runs. PSNR is mean paired clean/watermarked RGB PSNR over 500 SD-prompt
pairs; FID is from the separate 500-prompt COCO quality cohort.}}
\label{tab:alpha_sensitivity}
{\color{blue}\small\setlength{\tabcolsep}{3pt}
\begin{tabular}{ccccc}
\toprule
$\alpha$ & Accuracy (\%) $\uparrow$ & AUROC $\uparrow$ & PSNR (dB) $\uparrow$ & FID $\downarrow$ \\
\midrule
0.50 & $68.98\pm1.21$ & $0.7694\pm0.0090$ & 14.26 & 81.82 \\
0.75 & $66.27\pm0.71$ & $0.8312\pm0.0036$ & 14.42 & 82.52 \\
0.99 & $71.38\pm1.09$ & $0.7916\pm0.0106$ & 14.50 & 83.07 \\
\bottomrule
\end{tabular}}
\end{table}

\subsection{\textcolor{blue}{Sensitivity to Frequency Location}}
\label{subsec:frequency_location}
{\color{blue}
We test whether the centred low-frequency support used by the reference configuration is a
strong operating point or an arbitrary placement. The same 317-coefficient
spider-web support is placed at the low-frequency centre, at a conjugate-paired
offset of 12 frequency bins, and at a paired offset of 20 bins. Pattern values,
support capacity, $\alpha=0.99$, SD2.1, prompts, images, verifier architecture,
training schedule, nine evaluation conditions, five seeded evaluation runs,
and the fixed threshold $\tau=0.5$ are held constant. Each placement uses its
own matched verifier training because moving the support changes the detector
input distribution.

The centred setting achieves the highest observed fixed-threshold Accuracy,
71.38\% (Table~\ref{tab:frequency_location}). Moving to offset 12 reduces
Accuracy by 7.69 percentage points (95\% CI 5.82--9.56; Holm-adjusted
$p=0.00135$). Offset 20 achieves 70.16\% Accuracy and 0.8069 AUROC; neither
metric differs significantly from the centred setting in the four-test
location family. The results therefore support the centred reference support
as a strong and stable fixed-threshold choice, while showing that frequency
location affects performance. They do not establish a monotonic rule or a
unique theoretical optimum: the higher offset has slightly lower Accuracy but
slightly higher AUROC, and attack-specific effects differ by placement.
}

\begin{table}[t]
\centering
\caption{\textcolor{blue}{Frequency-location sensitivity with matched
verifier training. Values are macro mean $\pm$ SD over nine conditions and
five seeded evaluation runs.}}
\label{tab:frequency_location}
{\color{blue}\small\setlength{\tabcolsep}{3pt}
\begin{tabular}{lcc}
\toprule
Frequency location & Accuracy (\%) $\uparrow$ & AUROC $\uparrow$ \\
\midrule
Low-frequency centre & $71.38\pm1.09$ & $0.7916\pm0.0106$ \\
Offset 12 & $63.69\pm0.96$ & $0.7770\pm0.0118$ \\
Offset 20 & $70.16\pm0.70$ & $0.8069\pm0.0068$ \\
\bottomrule
\end{tabular}}
\end{table}

\subsection{\textcolor{blue}{Sensitivity to Verification Threshold}}
\label{subsec:threshold_sensitivity}
{\color{blue}
We vary only the final binary-decision threshold
$\tau\in\{0.3,0.4,0.5,0.6,0.7\}$ and reuse the saved predictions of the
reference verifier under all nine transformations. The checkpoint, images,
attacks, and five seeded evaluation runs are unchanged; no watermark
generation, DDIM inversion, or retraining is performed.

Macro Accuracy is stable from $\tau=0.3$ to 0.6, varying by at most 1.61
percentage points (Table~\ref{tab:threshold_sensitivity}). The highest observed
Accuracy occurs at $\tau=0.4$ (85.39\%), only 0.40 points above the current
$\tau=0.5$ setting (84.99\%). Retaining $\tau=0.5$ lowers FPR from 12.64\% to
8.89\% relative to 0.4 while preserving almost all of the observed Accuracy.
As expected, increasing $\tau$ reduces FPR and increases FNR. These results
show that SpiderMark is not dependent on a narrowly tuned threshold and
support $\tau=0.5$ as a balanced, pre-existing operating point rather than a
post-hoc Accuracy optimum.
}

\begin{table}[t]
\centering
\caption{\textcolor{blue}{Threshold sensitivity. All values are percentages
and macro mean $\pm$ SD over nine conditions and five evaluation runs.}}
\label{tab:threshold_sensitivity}
{\color{blue}\small\setlength{\tabcolsep}{3pt}
\begin{tabular}{lccc}
\toprule
$\tau$ & Accuracy $\uparrow$ & FPR $\downarrow$ & FNR $\downarrow$ \\
\midrule
0.3 & $85.29\pm0.96$ & $17.06\pm1.17$ & $12.60\pm1.46$ \\
0.4 & $85.39\pm1.10$ & $12.64\pm0.79$ & $16.37\pm1.92$ \\
0.5 (paper) & $84.99\pm0.69$ & $8.89\pm0.61$ & $20.51\pm1.32$ \\
0.6 & $83.78\pm0.50$ & $6.23\pm0.49$ & $25.20\pm0.90$ \\
0.7 & $81.27\pm0.44$ & $4.48\pm0.58$ & $31.53\pm0.70$ \\
\bottomrule
\end{tabular}}
\end{table}

\subsection{\textcolor{blue}{Inversion and Regeneration Attacks}}
\label{subsec:regeneration_attack}
{\color{blue}
SpiderMark is injected into the initial latent noise and propagated through
the diffusion process (Fig.~\ref{fig:spidernet}). An attacker with
access to a compatible diffusion model may therefore attempt to disrupt the
watermark by inverting a generated image, perturbing its recovered terminal
latent, and regenerating the image. To evaluate this diffusion-specific
removal threat, we use 50-step DDIM inversion and the same frozen SD2.1
model. Specifically,
\begin{equation}
x'_T=x_T+\sigma\epsilon,\qquad
\epsilon\sim\mathcal{N}(0,I),\quad
\sigma\in\{0.05,0.10,0.20\}.
\end{equation}
Inversion and 50-step regeneration use an empty prompt and guidance scale
1. Regenerated images are quantized to 8-bit RGB before a second 50-step
inversion and verification by the fixed reference verifier at threshold
0.5. Neither the diffusion model nor the verifier is retrained.
We use 150 validation images (79 watermarked and 71 clean) and five seeded
evaluation runs. Within each run, the three strengths share the same noise
direction for each image; the draw changes across runs. Original recovered
latents are cached and reused.
As shown in Table~\ref{tab:regeneration_attack}, Accuracy remains between
93.60\% and 94.93\%, while AUROC remains above 0.981. These results support
robustness to the evaluated moderate, non-adaptive
inversion--regeneration perturbations, with approximately 94\% overall
Accuracy without verifier adaptation. They do not establish resistance to
arbitrary regeneration or attackers optimizing against the verifier.
}

\begin{table*}[t]
\centering
\caption{\textcolor{blue}{Inversion/regeneration evaluation. Accuracy, FPR
and FNR are percentages; values are mean $\pm$ SD over five evaluation runs
on the same 150 images. Original images are unchanged across runs, hence
zero SD.}}
\label{tab:regeneration_attack}
{\color{blue}\small
\begin{tabular}{lcccc}
\toprule
Condition & Accuracy (\%) $\uparrow$ & AUROC $\uparrow$ & FPR (\%) $\downarrow$ & FNR (\%) $\downarrow$ \\
\midrule
Original & $94.67\pm0.00$ & $0.9902\pm0.0000$ & $9.86\pm0.00$ & $1.27\pm0.00$ \\
$\sigma=0.05$ & $93.60\pm1.01$ & $0.9817\pm0.0033$ & $7.89\pm0.77$ & $5.06\pm1.55$ \\
$\sigma=0.10$ & $93.87\pm1.19$ & $0.9824\pm0.0046$ & $6.76\pm1.18$ & $5.57\pm1.44$ \\
$\sigma=0.20$ & $94.93\pm1.21$ & $0.9848\pm0.0056$ & $4.79\pm1.61$ & $5.32\pm1.88$ \\
\bottomrule
\end{tabular}}
\end{table*}

\subsection{\textcolor{blue}{Generalisation to Stable Diffusion XL}}
\label{subsec:sdxl_generalisation}
{\color{blue}
We evaluate whether the SpiderMark construction extends beyond the SD2.1
backbone used in the main experiments. We generate 500 clean and 500
watermarked images with SDXL Base at $512\times512$, retaining the same
spider-web design, centred low-frequency support, $\alpha=0.99$, prompt and
seed rules, 50 inference steps, and guidance scale 7.5. The diffusion model
weights remain frozen. We evaluate nine conditions over five seeded runs.

The SD2.1 verifier applied directly to SDXL obtains 52.12\% macro Accuracy
and 0.7200 AUROC (Table~\ref{tab:sdxl_generalisation}). The above-chance
AUROC indicates that detectable watermark information remains, although the
fixed decision rule does not transfer reliably across generators. Training
only the same lightweight verifier on cached SDXL features raises performance
to 76.13\% Accuracy and 0.8481 AUROC, gains of 24.01 percentage points and
0.1281, respectively. This adaptation takes 383.5 seconds (approximately
6.4 minutes) on an RTX 4090 and does not retrain or fine-tune SDXL.

These results demonstrate practical cross-architecture portability: the
same injection construction operates on SDXL while only the lightweight
verification stage is adapted. They do not support architecture-independent
zero-shot verification; the learned verifier remains partly generator-specific.
}

\begin{table}[t]
\centering
\caption{\textcolor{blue}{Cross-architecture evaluation over nine conditions.
Values are macro mean $\pm$ SD over five seeded evaluation runs.}}
\label{tab:sdxl_generalisation}
{\color{blue}\small\setlength{\tabcolsep}{3pt}
\begin{tabular}{lcc}
\toprule
Generator and verifier & Accuracy (\%) $\uparrow$ & AUROC $\uparrow$ \\
\midrule
SD2.1 + reference verifier & $84.95\pm0.79$ & $0.9286\pm0.0067$ \\
SDXL + SD2.1 verifier & $52.12\pm0.48$ & $0.7200\pm0.0086$ \\
SDXL + adapted verifier & $76.13\pm1.13$ & $0.8481\pm0.0092$ \\
\bottomrule
\end{tabular}}
\end{table}

\subsection{\textcolor{blue}{Robustness under Stronger Transformations}}
\label{subsec:strong_transformations}
{\color{blue}
We further stress-test the fixed SD2.1 reference verifier at threshold 0.5
using the same 150-image validation split over five seeded evaluation runs.
The suite contains strong photometric change; fixed $30^\circ$ rotation;
perspective distortion with scale 0.50; random crops retaining only 25--40\%
of the image; and a deliberately extreme compound attack combining crop,
rotation, perspective distortion, photometric change, and JPEG quality 50.

SpiderMark retains 91.87\% Accuracy under strong photometric change, 85.33\%
under $30^\circ$ rotation, and 83.07\% under perspective distortion, with
AUROC above 0.90 in all three cases
(Table~\ref{tab:strong_transformations}). These results provide evidence of
strong resilience to severe individual photometric and geometric changes.
The severe crop is more challenging (66.13\% Accuracy and 0.8361 AUROC),
while the deliberately extreme compound attack identifies a clear robustness
boundary (50.93\% Accuracy and 0.6502 AUROC). Thus, the evidence supports
robustness to strong individual transformations, but not invariance to an
arbitrary composition of multiple severe transformations. Augmentation with
similarly compounded distortions is a promising extension.
}

\begin{table}[t]
\centering
\caption{\textcolor{blue}{Stronger-transformation evaluation with the
fixed reference verifier. Values are mean $\pm$ SD over five runs.}}
\label{tab:strong_transformations}
{\color{blue}\small\setlength{\tabcolsep}{3pt}
\begin{tabular}{lcc}
\toprule
Condition & Accuracy (\%) $\uparrow$ & AUROC $\uparrow$ \\
\midrule
Clean & $94.67\pm0.00$ & $0.9902\pm0.0000$ \\
Strong photometric & $91.87\pm0.73$ & $0.9760\pm0.0038$ \\
Rotation $30^\circ$ & $85.33\pm0.00$ & $0.9622\pm0.0000$ \\
Perspective 0.50 & $83.07\pm2.52$ & $0.9037\pm0.0107$ \\
Severe crop & $66.13\pm1.45$ & $0.8361\pm0.0248$ \\
Extreme compound & $50.93\pm1.80$ & $0.6502\pm0.0187$ \\
\bottomrule
\end{tabular}}
\end{table}

\subsection{\textcolor{blue}{Image-Frequency Removal Attacks}}
\label{subsec:frequency_removal}
\begin{figure*}[t]
\centering
\includegraphics[width=\textwidth]{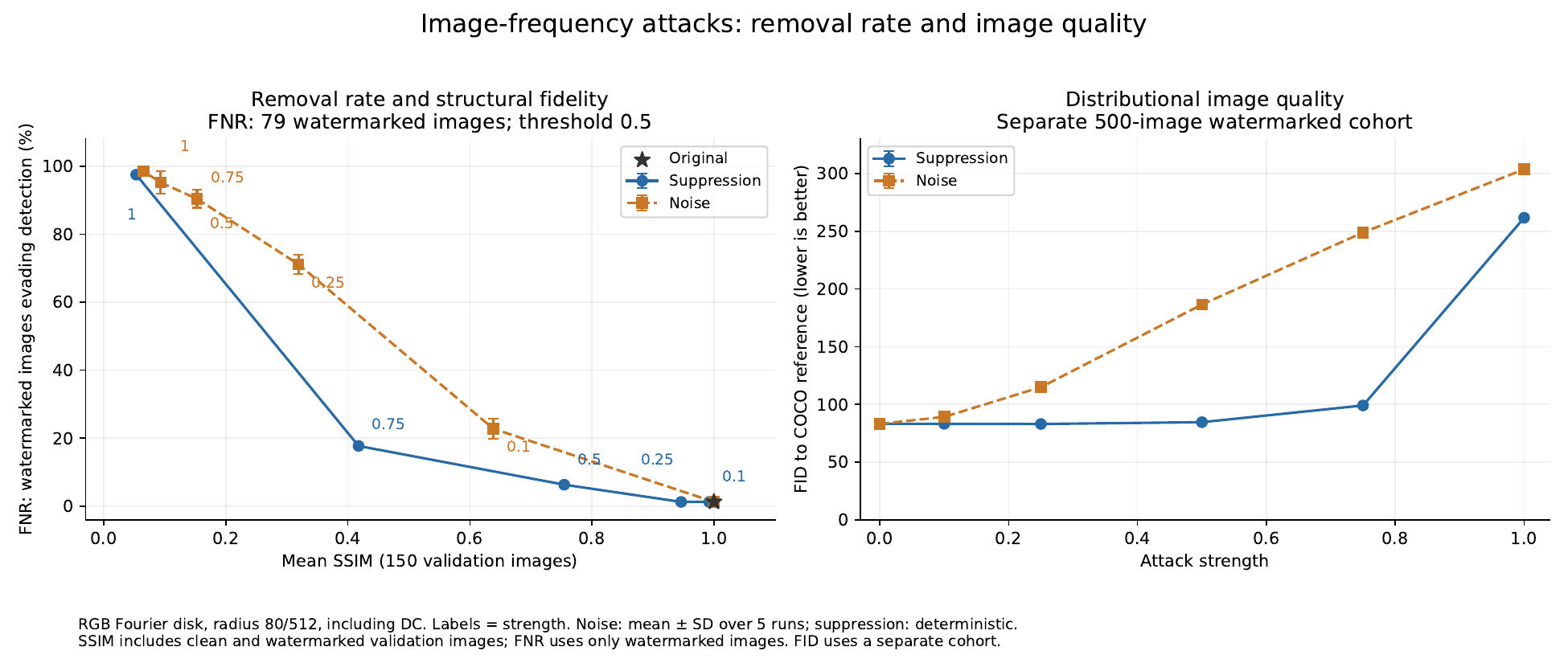}
\caption{\textcolor{blue}{Image-frequency removal and quality. Left: FNR
on 79 watermarked validation images versus mean SSIM on all 150 validation
images. Labels denote attack strength. Right: FID on a separate 500-image
watermarked COCO-prompt cohort. Noise error bars show SD across five runs;
suppression is deterministic. The RGB Fourier disk includes DC.}}
\label{fig:removal_fnr_tradeoff}
\end{figure*}

\begin{figure*}[t]
\centering
\includegraphics[width=\textwidth]{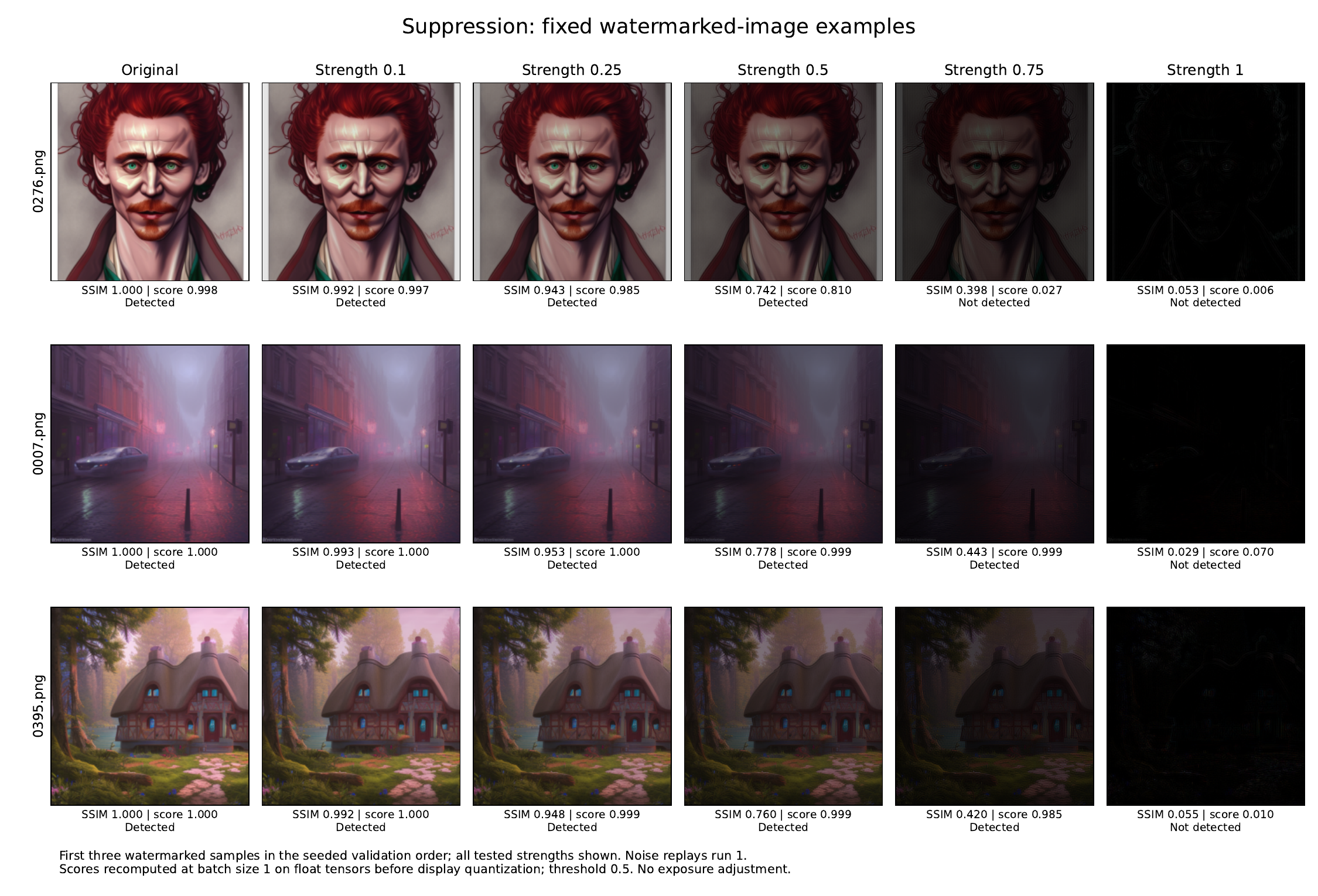}
\caption{\textcolor{blue}{Suppression examples: the first three watermarked
images in seed-42 validation order, with every tested strength shown.
Scores are recomputed at batch size one on float tensors before display
quantization; detection threshold is 0.5. No exposure adjustment is applied.
Darkening partly reflects DC removal. Some individual misses precede the
most severe degradation. Full noise examples appear in the supplementary material.}}
\label{fig:removal_examples}
\end{figure*}

{\color{blue}
The public, fixed watermark geometry creates a potential attack surface.
We evaluate suppression and noise within a complete RGB-image Fourier disk
of radius 80 at $512\times512$ resolution. This radius is scaled from the
latent radius 10 at resolution 64; it does not establish an exact mapping
of the latent watermark support through the diffusion decoder. These are
image-frequency attacks, not direct modifications of the latent spider-web.
For suppression, the masked coefficients are multiplied by $1-\lambda$.
For noise, masked Fourier noise is scaled to $\lambda$ times the masked
image-spectrum RMS. We test $\lambda\in\{0.1,0.25,0.5,0.75,1.0\}$ and clip
inverse-transformed images to $[0,1]$. The disk includes the DC component.

We use the existing verifier at threshold 0.5 with 50-step inversion.
Removal is measured by the false-negative rate (FNR), the proportion of
watermarked images that evade detection, rather than interpreting a drop
in overall Accuracy as removal success. FNR uses the 79 watermarked images
in the 150-image validation cohort; SSIM is averaged over all 150 images.
FID is evaluated separately on 500 watermarked COCO-prompt images.
Suppression is deterministic; noise results average five seeded runs.

Full suppression yields an FNR of 97.47\%, compared with 1.27\% on original
images, while SSIM decreases to 0.053. In the separate quality cohort, FID
increases from 83.14 to 261.68. Noise at $\lambda=0.5$ yields an FNR of
90.38\%, SSIM of 0.153 and FID of 186.55
(Figure~\ref{fig:removal_fnr_tradeoff}). Thus, near-complete removal under
these evaluated attacks accompanies substantial image degradation.
Individual images can nevertheless evade detection at weaker strengths
while retaining recognizable content (Figure~\ref{fig:removal_examples}).
To separate removal from the brightness loss caused by suppressing DC, a
full-cohort supplementary variant preserves the center bin. Even at
$\lambda=1.0$, 75 of 79 watermarked images remain detected (FNR 5.06\%),
while mean SSIM falls to 0.670. Its 58.00\% overall Accuracy is driven mainly
by an 83.10\% FPR rather than successful watermark removal.
These observations do not establish security against all adaptive attacks.
Keyed or randomized support and inversion-aware adaptive attacks remain
directions for future work. The supplementary material provides full
example grids and the full 150-image DC-preserving analysis.
}

\subsection{\textcolor{blue}{Sensitivity to Spider-Web Geometry}}
\label{subsec:geometry_sensitivity}
\begin{figure*}[t]
\centering
\includegraphics[width=\textwidth]{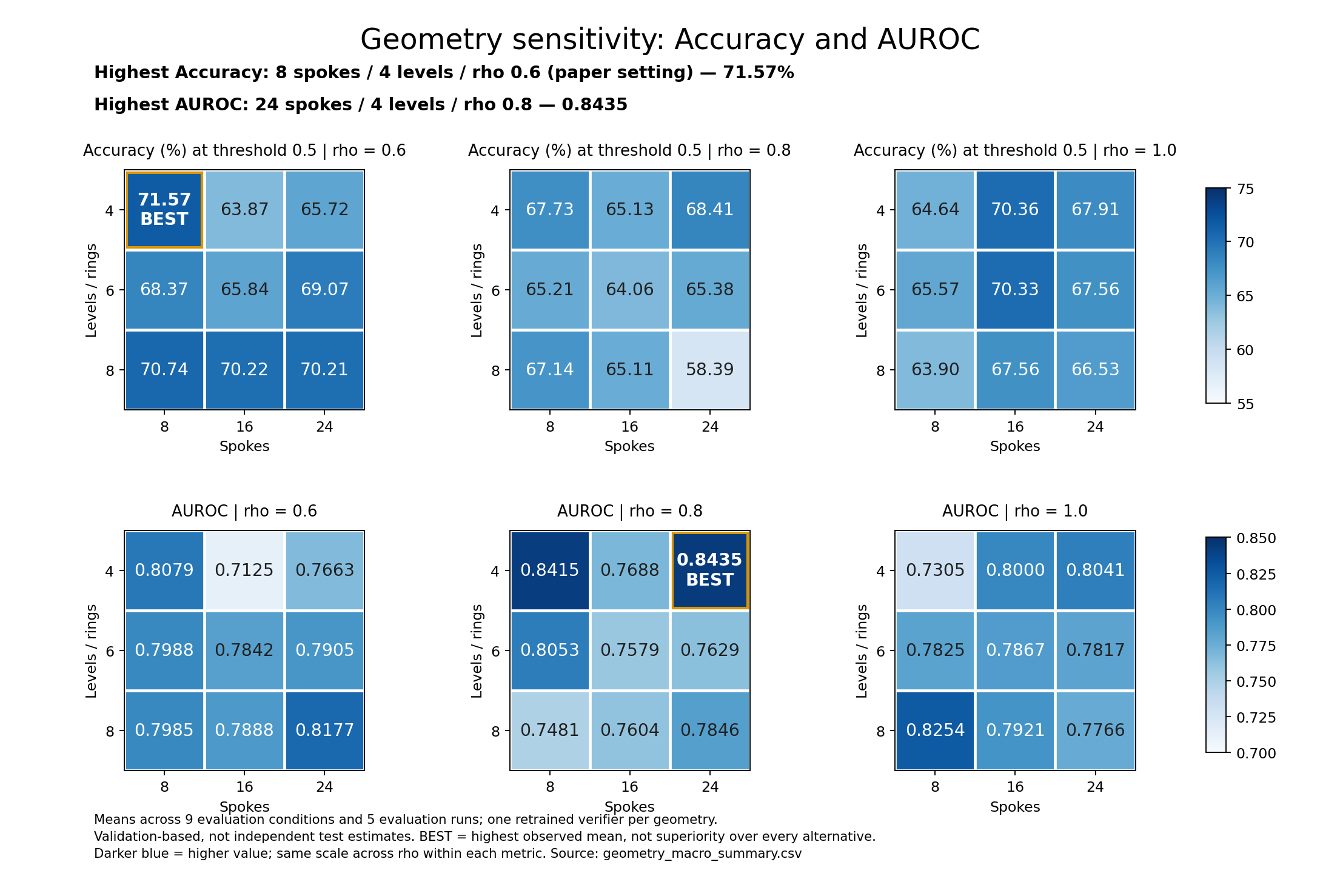}
\caption{\textcolor{blue}{Macro Accuracy at $\tau=0.5$ (top) and AUROC
(bottom) across all 27 geometry configurations. Columns within each panel
vary spokes; rows vary ring levels; panels vary $\rho$. Each metric shares
a color scale across panels. Gold outlines mark the highest observed mean.
Results use one retrained verifier per configuration and the validation-based
protocol described in Section~\ref{subsec:geometry_sensitivity}.}}
\label{fig:geometry_sensitivity}
\end{figure*}

{\color{blue}
We evaluated all 27 combinations of $n_s\in\{8,16,24\}$ spokes,
$n_\ell\in\{4,6,8\}$ ring levels, and $\rho\in\{0.6,0.8,1.0\}$.
The Stable Diffusion 2.1 Base backbone, prompts, generation seed indices,
embedding strength $\alpha=0.99$, and centred radius-10 Fourier injection
support were held fixed. The radius ratio changes the extent of the carrier
web, not the size of the injection support; only coefficients within that
support are replaced.

Each configuration used 500 clean and 500 watermarked images, with an
85\%/15\% training/validation split. One verifier was trained per geometry
using five cached augmentation rounds and 60 training epochs over that cache.
This controlled protocol differs from the reference checkpoint's
on-the-fly augmentation protocol, so its absolute scores should not be
equated with the main comparison results. Because deployment uses one
pre-specified operating point, fixed-threshold Accuracy at $\tau=0.5$ is the
primary metric in this sensitivity study; AUROC is reported as a complementary
threshold-independent ranking metric. Both are computed across nine evaluation
conditions, including clean images. Means and sample standard deviations
are computed across five seeded evaluation runs, not five independent
verifier training runs.

Table~\ref{tab:geometry_sensitivity} highlights the configurations with the
highest observed mean Accuracy and AUROC; Figure~\ref{fig:geometry_sensitivity}
shows all 27 settings. The original geometry $(8,4,0.6)$ achieved the highest
fixed-threshold macro Accuracy, whereas $(24,4,0.8)$ achieved higher AUROC
but lower Accuracy at the same operating threshold. Thus, the original
configuration is supported as a reasonable fixed-threshold operating point,
not as a universal optimum. More spokes or ring levels did not produce
consistent monotonic improvements. Across individual configurations,
Accuracy ranged from 58.39\% to 71.57\% and AUROC from 0.7125 to 0.8435;
small marginal factor effects do not imply stability across all combinations.

These findings are exploratory: checkpoints were selected using clean
validation loss, and the same validation images were subsequently evaluated
under the different transformations. They are therefore not independent
test-set estimates. The reported variability does not capture training-seed
uncertainty. Geometry-specific perceptual-quality comparisons and an
untouched test-set evaluation remain to be established. This factorial also
does not provide a matched isotropic-ring control.
}

\begin{table}[t]
\centering
\caption{\textcolor{blue}{Geometry sensitivity under controlled retraining.
Mean $\pm$ SD across five evaluation runs. Bold values denote the highest
observed mean among the 27 tested configurations for each metric, not
statistical superiority over every alternative.}}
\label{tab:geometry_sensitivity}
{\color{blue}\small
\begin{tabular}{ccc cc}
\toprule
$n_s$ & $n_\ell$ & $\rho$ & Accuracy (\%) $\uparrow$ & AUROC $\uparrow$ \\
\midrule
8 & 4 & 0.6 & $\mathbf{71.57}\pm0.85$ & $0.8079\pm0.0067$ \\
24 & 4 & 0.8 & $68.41\pm0.63$ & $\mathbf{0.8435}\pm0.0060$ \\
\bottomrule
\end{tabular}}
\end{table}

\subsection{\textcolor{blue}{Computational Cost of Verification}}
\label{subsec:verification_cost}
{\color{blue}
We compare online verification cost on an NVIDIA RTX 4090 using the same
Stable Diffusion 2.1 Base backbone, 50-step inversion, batch size one, and
30 images per method. Table~\ref{tab:verification_cost} reports mean latency,
the 95th-percentile latency (p95), and peak PyTorch-allocated GPU memory in
mebibytes (MiB). SpiderMark, Gaussian Shading~\cite{yang2024gaussian}, and
ROBIN~\cite{huang2024robin} have similar end-to-end costs: 1.154, 1.181,
and 1.174 seconds per image, respectively. SpiderMark has the lowest observed
mean latency, although the small differences do not establish a general speed
advantage. Its peak allocation is 2917.3 MiB, approximately 43--44 MiB above
the two baselines.

In a separate component-wise SpiderMark profile, recurring watermark
injection took 0.328 ms/image, 50-step DDIM inversion 1133.018 ms/image,
FFT/similarity extraction 1.036 ms/image, and verifier inference
2.010 ms/image on average. Injection excludes fixed-pattern construction.
These stage-wise measurements use a separate timing protocol and are not
summed to obtain the controlled whole-call comparison in
Table~\ref{tab:verification_cost}.
}

\begin{table}[t]
\centering
\caption{\textcolor{blue}{Controlled online verification cost on an RTX 4090
(30 images per method; batch size one). Memory is peak PyTorch allocation.
Latency includes preprocessing through detection but excludes disk I/O and
model loading.}}
\label{tab:verification_cost}
{\color{blue}\setlength{\tabcolsep}{2pt}\small
\begin{tabular}{lccc}
\toprule
Method & Mean $\pm$ SD (s) $\downarrow$ & p95 (s) $\downarrow$ & Memory (MiB) $\downarrow$ \\
\midrule
SpiderMark & $1.154\pm0.030$ & 1.192 & 2917.3 \\
Gaussian Shading & $1.181\pm0.033$ & 1.236 & 2873.6 \\
ROBIN & $1.174\pm0.035$ & 1.204 & 2874.7 \\
\bottomrule
\end{tabular}}
\end{table}

\section{Discussion}

This section synthesises the experimental findings and discusses their implications for practical watermark verification in diffusion-based generative models, with particular emphasis on robustness, generalisation, and deployment-oriented evaluation.

\subsection{Overall Robustness of the Learned Detector}

The results consistently demonstrate that SpiderMark achieves strong and stable watermark verification across a wide range of photometric, geometric, and mixed distortions. Unlike other watermarking methods, the learned verifier maintains reliable performance even under aggressive transformations such as strong JPEG compression, spatial resampling, and partial occlusion.

{\color{blue}The stronger stress test in Section~\ref{subsec:strong_transformations}
reinforces this conclusion for individual severe distortions: Accuracy remains
between 83.07\% and 91.87\% under strong photometric change, $30^\circ$
rotation, and perspective distortion, with AUROC above 0.90. Severe crop and
the deliberately extreme compound condition identify the boundary of this
robustness rather than diminishing the strong individual-attack results.}

{\color{blue}The detector also retains useful performance under the tested
misalignment and local corruption. This is consistent with the injected
frequency structure persisting through diffusion and downstream processing,
although the experiments do not establish invariance to arbitrary spatial
transformations or perceptual indistinguishability.}

\subsection{Limitations of AUC as a Robustness Metric}\label{subsec:auc_limitations}

\begin{figure*}
    \centering
    \includegraphics[width=1\linewidth]{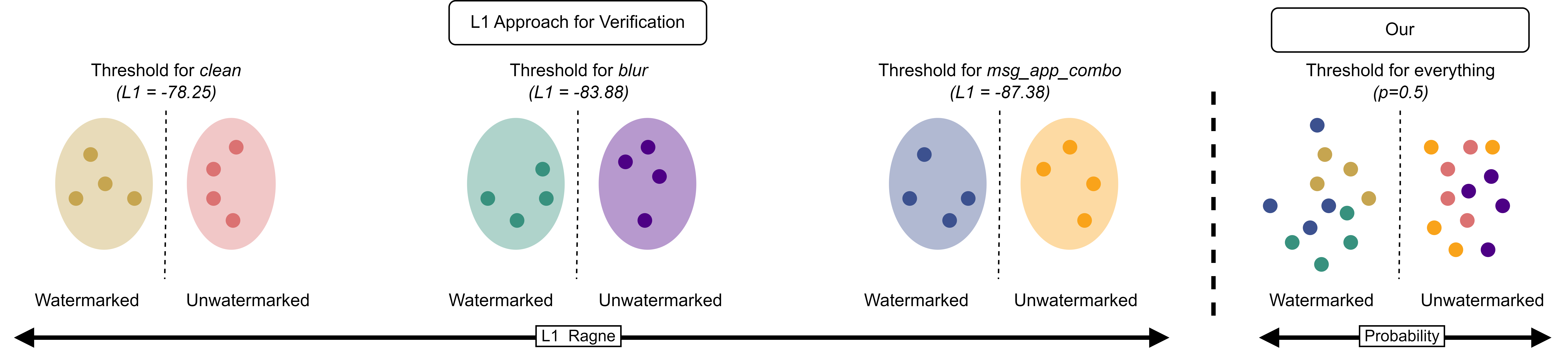}
    \caption{Conceptual comparison between L1-distance--based verification and our probabilistic verification approach. 
        \textbf{Left:} \textcolor{blue}{L1-distance distributions shift across attacks, making a shared decision threshold harder to select.}
        \textbf{Right:} Our method maps verification evidence to \textcolor{blue}{a learned continuous confidence score}, enabling \textcolor{blue}{fixed-threshold decisions across the evaluated attacks without claiming attack invariance}.}
    \label{fig:l1_limitations}
\end{figure*}

\begin{table}[h!]
\centering
\caption{Optimal thresholds for L1 and PSNR detectors under different attacks. Large threshold shifts indicate that AUC alone is unreliable for evaluating watermark detection robustness.}
\label{tab:threshold_shift}
\begin{tabular}{lcc}
\hline
\textbf{Attack} & \textbf{Best L1 Thr} & \textbf{Best PSNR Thr} \\
\hline
clean          & -78.2500  & 5.3937 \\
jpeg\_strong   & -82.2500  & 4.9414 \\
msg\_app\_combo & -87.3750  & 4.3452 \\
down\_up       & -82.5625  & 4.6217 \\
blur           & -83.8750  & 4.0488 \\
random\_crop   & -82.6875  & 5.0281 \\
occlusion      & -79.9375  & 5.2857 \\
geom\_warp     & -81.0625  & 5.2801 \\
train\_aug\_mix & -82.1875 & 4.9464 \\
\hline
\end{tabular}
\end{table}

Although AUROC is a common metric for evaluating binary classifiers, our experiments highlight its limitations for assessing watermark robustness in real-world deployment scenarios. In practice, watermark verification systems must operate at a \emph{fixed decision threshold}, whereas AUROC evaluates separability by sweeping over all possible thresholds. As a result, a high AUROC does not necessarily translate into reliable verification performance at inference time.

{\color{blue}As shown in Table~\ref{tab:threshold_shift}, the empirically
best thresholds for standalone L1 and PSNR detectors differ across attacks.
Thus, good ranking performance alone need not imply reliable detection at a
single preselected operating threshold.}

{\color{blue}Figure~\ref{fig:l1_limitations} illustrates the attack-dependent
score shifts. SpiderMark instead combines spectral evidence and auxiliary
cues in a learned continuous score, evaluated at the same fixed threshold
across conditions. This is more directly aligned with deployment than
selecting a separate threshold after observing each attack; we do not claim
formal probability calibration.}

\subsection{Photometric Robustness}

As reported in Table~\ref{tab:robustness_results_all}, SpiderMark retains 87.20\% Accuracy
under strong JPEG compression and 81.60\% under Gaussian blur. In the stronger
photometric stress test (Section~\ref{subsec:strong_transformations}), it
achieves 91.87\% Accuracy and 0.9760 AUROC. These results show that the learned
verifier remains effective after substantial quantisation and smoothing.

This robustness is particularly important in real-world content distribution pipelines, where images are routinely re-encoded, resized, or compressed by platforms beyond the control of the content creator or verifier.

\subsection{Geometric and Spatial Robustness}

Spatial transformations pose a harder challenge because they alter the
watermark's spatial--frequency alignment. Nevertheless, Table~\ref{tab:robustness_results_all}
reports 91.33\% Accuracy after downsampling--upsampling, 87.87\% after random
cropping, and 85.60\% after geometric warping. Under the stronger tests in
Section~\ref{subsec:strong_transformations}, SpiderMark achieves 85.33\%
Accuracy for a $30^\circ$ rotation and 83.07\% for perspective distortion.
These results support robustness under the evaluated geometric transforms;
we do not claim general geometry invariance.

\subsection{\textcolor{blue}{Contribution of the Verifier and Auxiliary Cues}}

{\color{blue}Section~\ref{subsec:verifier_ablation} shows that the learned
verifier improves detection over the tested classical correlation detector
while keeping SpiderMark injection fixed. Within the controlled retraining
protocol, adding the PSNR--L1 cue pair to spectral features increases macro
Accuracy from 67.88\% to 72.86\%. This comparison establishes a joint cue
contribution, not separate benefits from PSNR and L1. It does not remove the
injected watermark or its support mask, and therefore cannot establish the
superiority of the spider-web pattern over other geometries.}

\subsection{\textcolor{blue}{Cross-Dataset and Cross-Architecture Generalisation}}

A key finding of this work is the strong generalisation of the verifier to unseen data distributions. Although the verifier is trained exclusively on images generated from the Stable Diffusion prompt dataset, it consistently achieves the best performance on the COCO dataset without any retraining or fine-tuning.

This result indicates that the verifier does not overfit to prompt-specific semantics or dataset-specific artefacts, but instead learns watermark-consistent frequency-domain cues that transfer across datasets. Such generalisation is essential for practical deployment, where the verifier may encounter images generated under diverse prompts, styles, or downstream usage contexts.

{\color{blue}The SDXL study in Section~\ref{subsec:sdxl_generalisation}
extends this evidence from dataset shift to generator shift. The unchanged
SpiderMark injection construction operates on SDXL, and adapting only the
lightweight verifier for approximately 6.4 minutes achieves 76.13\% Accuracy
and 0.8481 AUROC across nine conditions without modifying or fine-tuning the
diffusion model. This establishes practical portability while making clear
that the verifier is partly generator-specific.}

\subsection{Semantic Preservation and Perceptual Quality}

{\color{blue}CLIP image--text similarity differs from clean generations by
less than 0.003 on both prompt cohorts. Together with the numerically close
FID scores, this indicates that SpiderMark's stronger fixed-threshold
detection is accompanied by small changes in these measured quality and
alignment metrics. Neither metric alone proves perceptual equivalence.}

\subsection{\textcolor{blue}{Security Scope and Threat Model}}
\label{subsec:security_scope}
{\color{blue}
SpiderMark is designed for robust provenance verification under downstream
processing. We distinguish this objective from cryptographic authentication
against a fully informed adversary. Our operational threat model assumes a
trusted generator or watermark issuer and a verifier with the required
reference information. The attacker may transform or attempt to remove the
mark from an output image, but the current construction does not claim to
provide identity binding, non-repudiation, or secrecy of the fixed spider-web
support.

The current implementation provides binary presence detection rather than a
multi-bit payload. This is appropriate for determining whether an image was
produced by a protected generation pipeline, but it does not encode model
versions, issuer identifiers, or per-image metadata. Extending SpiderMark to
fine-grained tracing would require a keyed family of radial/angular patterns
or component modulation together with a dedicated decoder and evaluation of
payload capacity, bit-error rate, robustness, and security--capacity
trade-offs. We leave this payload-bearing extension to future work.

The informed removal experiment in Section~\ref{subsec:frequency_removal}
quantifies a practically favorable removal--utility trade-off. Complete
suppression ($\lambda=1.0$) raises the watermarked-image false-negative rate
to 97.47\%, but reduces SSIM to 0.053 and increases FID from 83.14 to 261.68.
Targeted noise at $\lambda=0.5$ yields a 90.38\% false-negative rate, together
with SSIM 0.153 and FID 186.55. Thus, near-complete evasion is possible for
an attacker who knows the support, but in these tests it requires severe
degradation that substantially reduces the utility of the resulting image.

The fixed public geometry nevertheless creates an attack surface beyond this
experiment. An adversary could attempt forgery by inserting a detectable
pattern into unrelated content or by optimizing against the verifier; the
present detector establishes watermark presence rather than cryptographically
binding an image to an issuer. Exposure of a future secret pattern would
similarly require key rotation and revocation. Collusion across multiple
marked outputs could help estimate shared structure, and a white-box or
query-based attacker could optimize removal or false acceptance directly.
These forgery, key-leakage, collusion and fully adaptive threats are not
resolved by the current experiments and are not covered by our robustness
claim.

Promising extensions include keyed and per-image randomized supports,
cryptographic binding of watermark identifiers to an issuer, key rotation,
query-aware evaluation, and adversarial training against adaptive removal and
forgery. This explicit scope prevents transformation robustness from being
misinterpreted as cryptographic security while preserving the empirical
finding that effective informed removal in the evaluated setting incurs a
large quality cost.
}

\subsection{Implications for Real-World Deployment}

Taken together, these findings establish SpiderMark as a \textcolor{blue}{practical watermarking framework} for diffusion-based generative models. By combining noise-level frequency injection with \textcolor{blue}{continuous, threshold-aware} verification, the proposed system addresses key limitations of existing watermarking approaches, including threshold instability, poor robustness under transformation, and limited generalisation.

The analysis also highlights the importance of evaluation protocols that reflect real-world operating constraints. \textcolor{blue}{Fixed-threshold metrics reported alongside continuous-score measures} provide a more faithful assessment of robustness than threshold-free measures such as AUROC alone. These insights are broadly applicable beyond watermarking and may inform the evaluation of other content attribution and authenticity systems.

\section{Conclusion}

This paper introduced \textit{SpiderMark}, a structured watermarking framework for diffusion-based image generation that embeds a fixed geometric pattern directly into the noise initialization stage in the frequency domain. Unlike prior approaches that rely on handcrafted similarity metrics or rigid frequency ring structures, SpiderMark unifies watermark injection and probabilistic verification within a single framework that remains fully compatible with pretrained diffusion models and requires no modifications to model weights.

Extensive evaluations demonstrate that the proposed verifier substantially outperforms PSNR-based detection across a wide range of photometric, geometric, and mixed distortions. By learning watermark-consistent frequency cues rather than relying on pixel-level alignment, SpiderMark maintains reliable performance under compression, resampling, occlusion, and spatial warping. {\color{blue}With watermark injection held fixed, the ablations support the learned verifier and the joint contribution of its PSNR--L1 auxiliary cue pair; they do not assign separate gains to the two cues.}

{\color{blue}The reference verifier transfers to unseen COCO prompts without
retraining, and CLIP and FID remain numerically close to the corresponding
comparison images and methods. These results support cross-prompt transfer
and a favorable measured detection--quality balance, while stopping short
of claims of universal transfer or perceptual equivalence.}

{\color{blue}Beyond cross-dataset transfer, the same watermark construction
was deployed on SDXL without modifying its weights; adapting only the
lightweight verifier yielded 76.13\% Accuracy and 0.8481 AUROC. Under stronger
individual transformations, SpiderMark retained 83--92\% Accuracy and AUROC
above 0.90 for strong photometric, rotation, and perspective conditions.}

Beyond empirical performance, this work highlights important evaluation considerations for practical watermarking systems. In particular, we demonstrate that threshold-free metrics such as AUROC can be misleading under deployment constraints, and that \textcolor{blue}{continuous-score, threshold-aware verification} provides a more reliable measure of real-world robustness.

Overall, SpiderMark provides a practical and generalisable solution for content attribution in diffusion-based generative models. Future work may extend this framework to multimodal generation pipelines, investigate adaptive or content-aware watermarking strategies, and explore robustness against more extreme geometric or adversarial transformations.

{\color{blue}SpiderMark should therefore be interpreted as a robust provenance
verification method rather than a complete cryptographic authentication
scheme. Under the evaluated informed removal attacks, near-complete evasion
incurs severe image-quality degradation; keyed per-image supports,
issuer binding, collusion analysis, and adaptive security evaluation remain
important extensions.}

\bibliographystyle{IEEEtran}
\bibliography{main}

\end{document}